%% file: 0_Main.tex
\documentclass[10pt,journal,compsoc, twoside]{IEEEtran}
\ifCLASSOPTIONcompsoc
  \usepackage[nocompress]{cite}
\else
  \usepackage{cite}
\fi
\ifCLASSINFOpdf
\else
\fi
\usepackage{amsmath}
\usepackage{algorithmic}
\usepackage{array}
\ifCLASSOPTIONcompsoc
 \usepackage[caption=false,font=footnotesize,labelfont=sf,textfont=sf]{subfig}
\else
 \usepackage[caption=false,font=footnotesize]{subfig}
\fi
\usepackage{fixltx2e}
\usepackage{stfloats}
\ifCLASSOPTIONcaptionsoff
 \usepackage[nomarkers]{endfloat}
\let\MYoriglatexcaption\caption
\renewcommand{\caption}[2][\relax]{\MYoriglatexcaption[#2]{#2}}
\fi

\usepackage{multicol}
\usepackage{multirow}
\usepackage{hhline}
\usepackage{makecell}
\usepackage{bm}
\usepackage{enumerate}
\usepackage{color}
\usepackage{enumitem}
\usepackage{booktabs}
\usepackage{graphicx}
\usepackage{ulem}
\usepackage{stfloats}
\usepackage{verbatim}
\usepackage{amsmath}
\usepackage{algorithm}
\usepackage{algorithmic}
\usepackage{footnote}
\usepackage{url}
\usepackage{cite}
\usepackage{amsfonts}
\usepackage{etoolbox}
\usepackage{indentfirst}
\usepackage{epstopdf}
\usepackage{amsmath}
\usepackage{amssymb}
\usepackage{booktabs}
\usepackage{multirow}
\usepackage{colortbl}
\usepackage{bbding}
\usepackage[table,xcdraw]{xcolor}
\begin{document}
\title{Voice-Face Homogeneity Tells Deepfake}

\author{
      Harry~Cheng,~\IEEEmembership{Student Member,~IEEE},
      Yangyang~Guo~\IEEEmembership{Member,~IEEE},
      Tianyi~Wang,~\IEEEmembership{}
      Qi~Li,~\IEEEmembership{}
      Xiaojun~Chang,~\IEEEmembership{Senior Member,~IEEE}
      Liqiang~Nie,~\IEEEmembership{Senior Member,~IEEE}
\IEEEcompsocitemizethanks{
\IEEEcompsocthanksitem Harry Cheng and Qi~Li are with the School of Computer Science and Technology, Shandong University, Qingdao, China, e-mail: \{xaCheng1996, iliqi.nice\}@gmail.com.

\IEEEcompsocthanksitem Yangyang Guo is with the School of Computing, National University of Singapore, Singapore, e-mail: guoyang.eric@gmail.com.

\IEEEcompsocthanksitem Tianyi Wang is with the Department of Computer Science, The University of Hong Kong, HongKong, China, e-mail: tywang@cs.hku.hk.

\IEEEcompsocthanksitem Xiaojun Chang is with Faculty of Engineering and Information Technology, University of Technology Sydney,  Sydney, Australia, e-mail: cxj273@gmail.com.

\IEEEcompsocthanksitem Liqiang Nie is with the Department of Computer Science and Technology, Harbin Institute of Technology (Shenzhen), Shenzhen, China, e-mail: nieliqiang@gmail.com.}% 

\thanks{
Corresponding authors: Yangyang Guo and Liqiang Nie.
}
}

% \markboth{IEEE Transactions on Pattern Analysis and Machine Intelligence}{Cheng et al.: Voice-Face Homogeneity Tells Deepfake}

\IEEEtitleabstractindextext{%
\begin{abstract}
   Detecting forgery videos is highly desirable due to the abuse of deepfake. Existing detection approaches contribute to exploring the specific artifacts in deepfake videos and fit well on certain data. However, the growing technique on these artifacts keeps challenging the robustness of traditional deepfake detectors. As a result, the development of generalizability of these approaches has reached a blockage. To address this issue, given the empirical results that the identities behind voices and faces are often mismatched in deepfake videos, and the voices and faces have homogeneity to some extent, in this paper, we propose to perform the deepfake detection from an unexplored voice-face matching view. To this end, a voice-face matching method is devised to measure the matching degree of these two. Nevertheless, training on specific deepfake datasets makes the model overfit certain traits of deepfake algorithms. We instead, advocate a method that quickly adapts to untapped forgery, with a pre-training then fine-tuning paradigm. Specifically, we first pre-train the model on a generic audio-visual dataset, followed by the fine-tuning on downstream deepfake data. We conduct extensive experiments over three widely exploited deepfake datasets - DFDC, FakeAVCeleb, and DeepfakeTIMIT. Our method obtains significant performance gains as compared to other state-of-the-art competitors. It is also worth noting that our method already achieves competitive results when fine-tuned on limited deepfake data.
\end{abstract}

}

\maketitle

\IEEEdisplaynontitleabstractindextext

% \fi
% \IEEEpeerreview
\maketitle

\input{1_intro}
\input{2_rel}
\input{3_prem}
\input{4_exp}

\input{5_conclusion}

% or


% % use section* for acknowledgment
% \ifCLASSOPTIONcompsoc
%   % The Computer Society usually uses the plural form
%   \section*{Acknowledgments}
% \else
%   % regular IEEE prefers the singular form
% \fi


\ifCLASSOPTIONcaptionsoff
  \newpage
\fi



%\bibliographystyle{IEEEtran}

\normalem
\bibliographystyle{IEEEtran}
\bibliography{ieeebib}

% \input{6_author_bio}


% % insert where needed to balance the two columns on the last page with
% %\newpage



%\vfill


\end{document}

%% file: 1_intro.tex
\section{Introduction}
\label{sec:intro}
\IEEEPARstart{D}{eepfake}~\cite{df1,df2,df3,df4} is to synthesize the media in which a person is replaced with someone else's portrait or vocal. Given its successful application in animation~\cite{Tip_cite_1} and online education~\cite{df6}, deepfake has attracted increasing interest from academic and industrial practitioners. However, the abuse of such techniques, such as maliciously editing porn and violent videos\footnote{\url{https://www.bbc.com/news/technology-42912529}.}, seriously challenges social functioning and ethics, spawning widespread concerns. It is hence imperative to detect deepfake abuse with effective measures. 

\begin{figure}[t]
\centering
\subfloat[Trained on FF++, tested on FF++, AUC=99.04]{\includegraphics[width=0.45\textwidth]{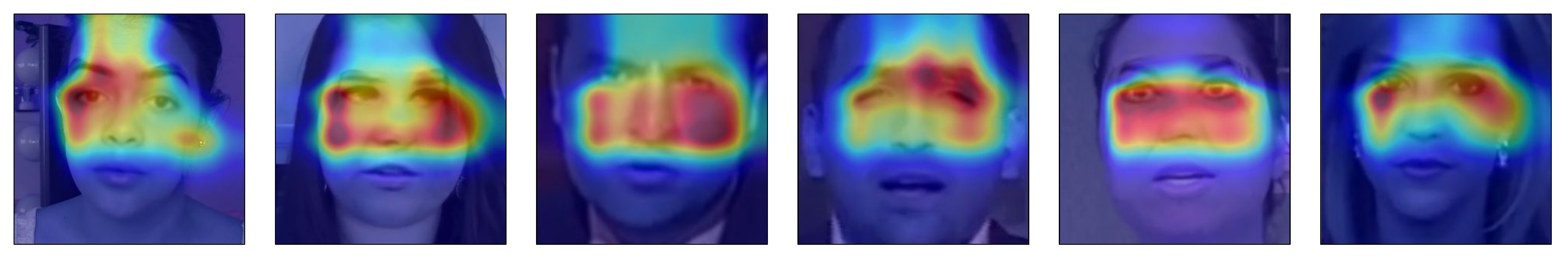}}
\hfill
\subfloat[Trained on FF++, tested on DFDC, AUC=60.51]{\includegraphics[width=0.45\textwidth]{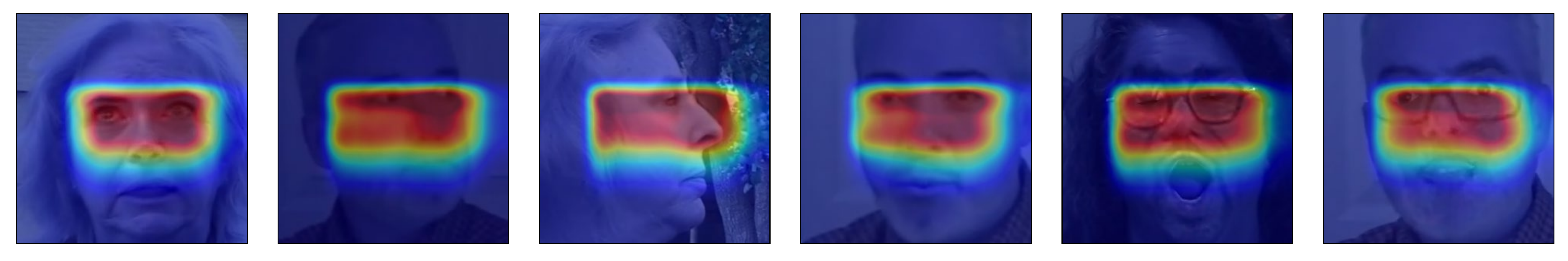}}
\hfil
\subfloat[Trained on DFDC, tested on DFDC, AUC=75.52]{\includegraphics[width=0.45\textwidth]{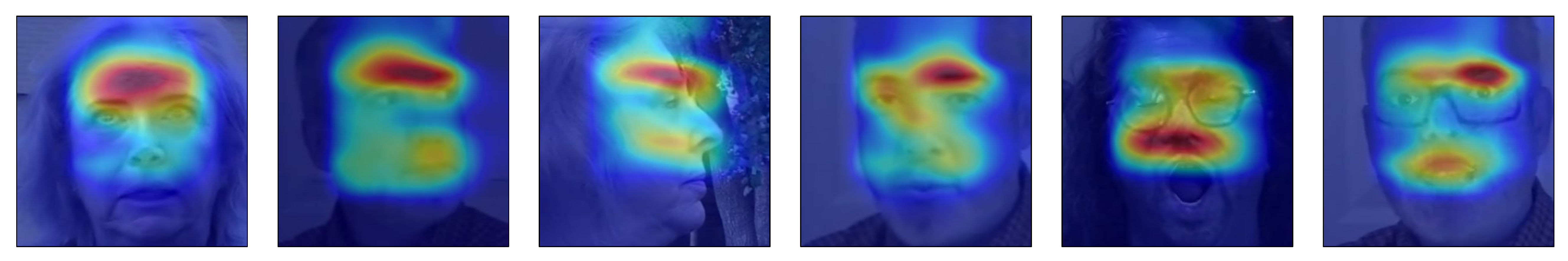}}
\hfil
\subfloat[Trained on DFDC, tested on FF++, AUC=57.43]{\includegraphics[width=0.45\textwidth]{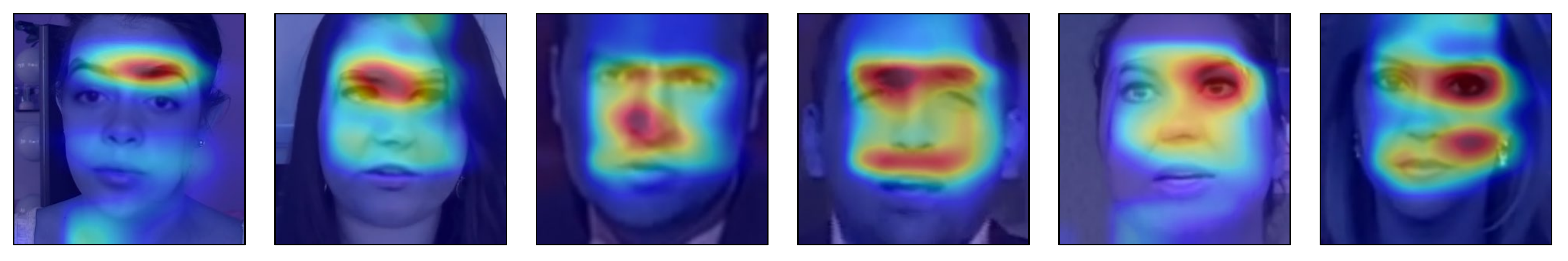}}
\hfil
\caption{Heat maps and AUC scores (\%) from Xception~\cite{dfd1} under different settings. When trained on FF++, the attention is mostly placed on eyes regions, and is shifted to the forehead and nose for DFDC. Since the editing algorithm focuses on different regions with respect to distinctive data, a well-trained model thus cannot adapt smoothly across datasets.}
\label{example}
\end{figure}

% \centering
% \subfloat[]{\includegraphics[width=2.5in]{fig1}%
% \label{fig_first_case}}
% \hfil
% \subfloat[]{\includegraphics[width=2.5in]{fig1}%
% \label{fig_second_case}}
Considerable research efforts have been dedicated to detecting deepfakes thus far~\cite{F3Net,dfd1,dfd4,DBLP:journals/tip/WangGZ22}. Most of them explore the face manipulation artifacts of fake videos, including the visual artifacts from face attentive regions~\cite{attm}, apparent changes in the frequency domain resulting from up-sampling~\cite{DBLP:conf/cvpr/LiuLZCH0ZY21}, or amplified artifacts via isolating manipulated faces~\cite{dfd3}. Nevertheless, these methods are all limited by one critical downside, namely, \textit{inferior generalization across datasets}. For instance, a model trained on the FF++ dataset~\cite{dfd1} suffers significant performance degradation when migrated to other deepfake datasets (\textit{e.g.}, DFDC~\cite{dfdc} or Celeb-DF~\cite{celeb-df}). The key reason is that different datasets are built with distinctive manipulation algorithms. As a result, previous detection approaches tend to fit well on the specific training data, and the generalization is thereby hampered. For instance, Figure~\ref{example} illustrates that the models trained on FF++ pay more attention to the eyes yet fail on DFDC since the salient regions are instead the forehead and nose. 

Existing solutions for this generally contribute from two directions. The first direction is to enhance the detection of manipulated traces with complementary modalities~\cite{emotion}. For instance, Zhou \textit{et al.}~\cite{jointAV} leverage the speech content to unearth unharmonious mouth-related dynamics. However, these approaches focus on partial facial characteristics, such as lip movements~\cite{jointAV} or emotional biases~\cite{emotion}, which are easily attacked by specialized countermeasures~\cite{df6}. The other direction is to extend current datasets with auxiliary visual features, \textit{e.g.}, blending regions~\cite{x-ray} or facial landmarks~\cite{lipsdontlie}. Since these features are not fully covered by existing forgery algorithms, methods from this direction make it easy to spot the salient differences between real and fake frames. Nevertheless, building such datasets is time-consuming, and the models often require computationally intensive training on both forgery and auxiliary data.
\begin{figure*}[t]
    \centering
    \subfloat[Pre-training Stage]{\includegraphics[width=0.4380\textwidth]{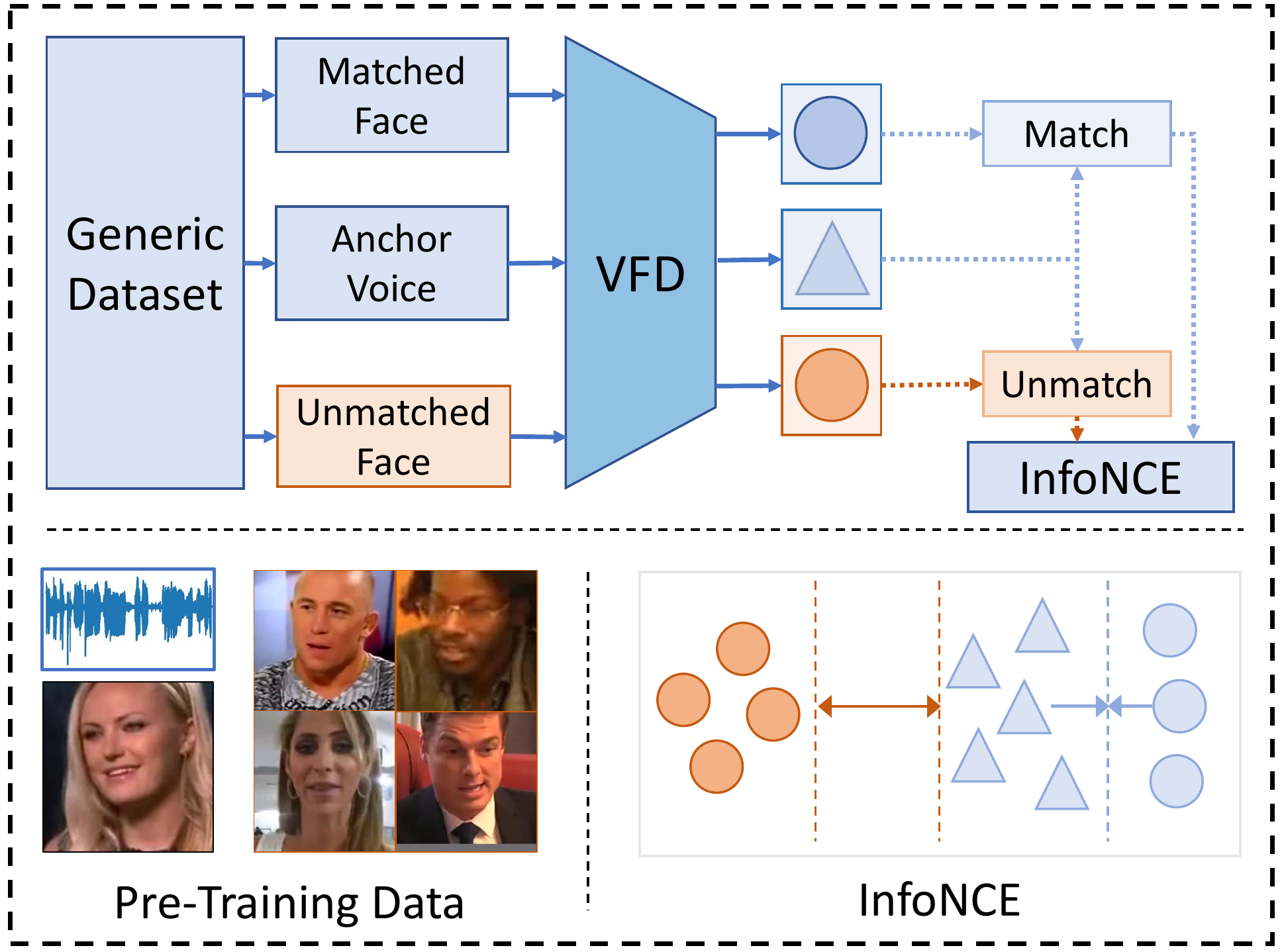}}
    \subfloat[Fine-tuning and testing Stage]{\includegraphics[width=0.50\textwidth]{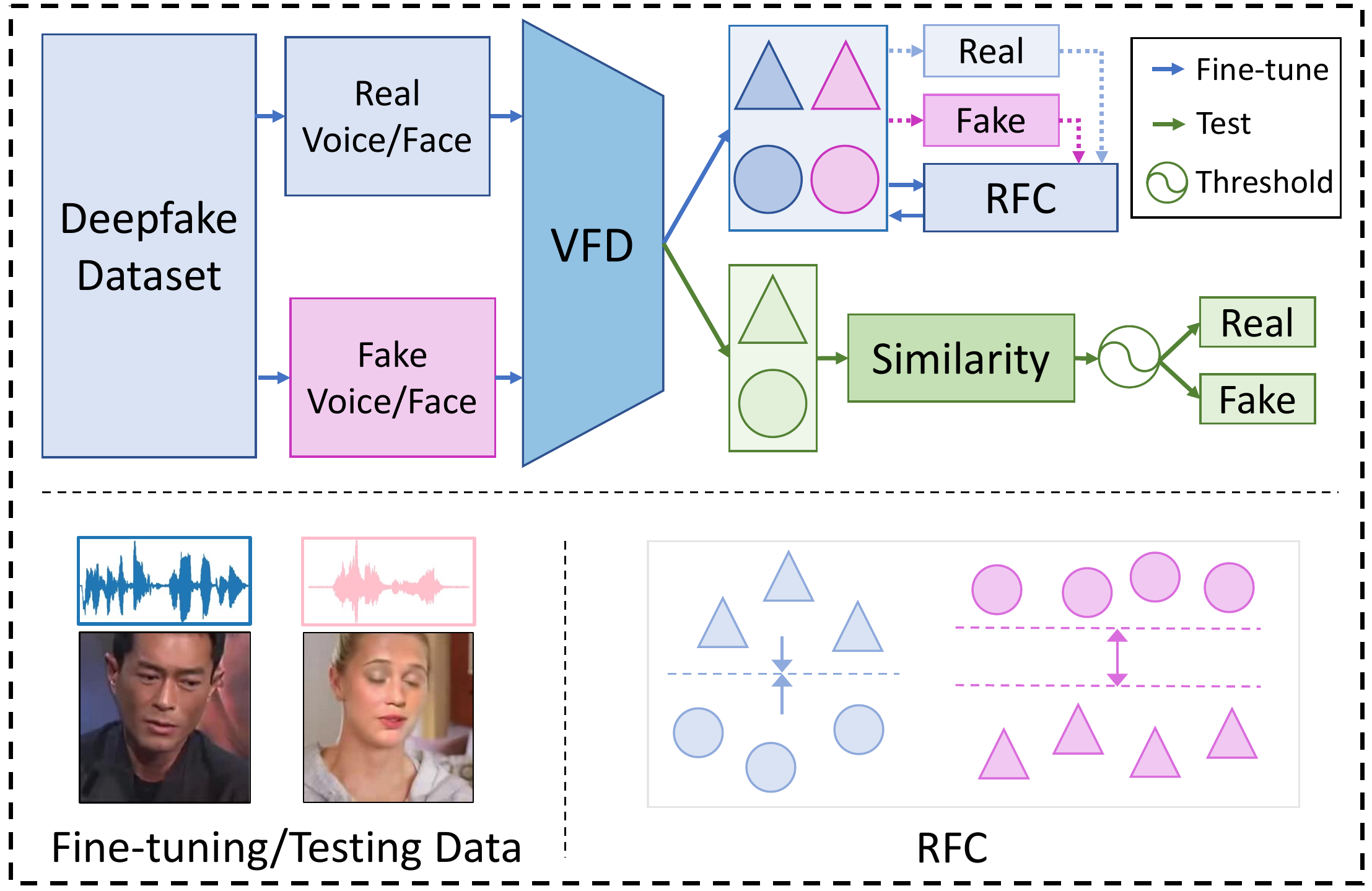}}
    \hfill
    \caption{Schematic illustration of our proposed VFD model. In the pre-training stage (a), we adopt the InfoNCE loss to pull matched voices (blue triangles) and faces (blue circles) closer while pushing unmatched faces (orange circles) faraway. For the fine-tuning (b), the upgraded real-fake contrastive loss is employed to fine-tune the model on the deepfake dataset. The cosine similarity serves as the evidence for judging the video authenticity during inference.}
    \label{overview} 
\end{figure*}

To overcome the shortcomings encountered by the existing studies, this paper devotes to addressing this task from an unexplored angle. Our solution is inspired by the recent progress from cross-modal biometric matching~\cite{DBLP:conf/cvpr/NagraniAZ18, DBLP:conf/cvpr/OhDKMFRM19}, which certifies the fact that an individual's voice and face are strongly correlated. Based upon this finding, we conduct an in-depth analysis of the existing deepfake algorithms and recognize a serious voice-face mismatching problem. For example, an edited face of President Obama can be matched with the voice of President Reagan. This motivates us to speculate - Can we perform deepfake detection from a discrimination view of voices and faces?

To answer this question, we tentatively propose a practical method called Voice-Face matching Detection, VFD for short. In particular, we perform matching between voices and faces rather than directly attacking the artifacts from deepfake. An overview of VFD is shown in Figure~\ref{overview}. We first train VFD on a generic audio-visual dataset (such as Voxceleb2~\cite{voxceleb2}) without being manipulated by deepfake algorithms. Specifically, given an anchor voice clip, the popular InfoNCE loss~\cite{InfoNCE} is employed to pull matched voice clip and face instances closer while pushing unmatched ones further. As the pre-training dataset contains rich data for building an informative feature extractor, our VFD is enabled to be easily adapted to various downstream deepfake datasets with few fine-tuning steps. On the basis of InfoNCE, we design an upgraded Real-Fake Contrastive loss function, termed as RFC, to align the objective between pre-training and fine-tuning, given that the downstream deepfake dataset contains real (positive) and fake (negative) pairs. Arguably, our method offers a best-of-both-worlds solution - 1) VFD focuses on the general matching objective of voices and faces and can be quickly migrated to various deepfake datasets, as opposed to paying attention to designated face regions (see Figure 1). 2) The pre-training then fine-tuning paradigm alleviates the requirement of auxiliary data. Our model builds upon easily collected generic videos and depends less on the expensive auxiliary data from deepfake. 

% Thereafter, instead of forcing the model to detect untapped forgery directly, we advocate for one that adapts to unprecedented forgery. In particular, the trained model is smoothly transferred to fine-tune on the forgery dataset to narrow the gap between generic and deepfake videos. Due to the coarse annotation, it is non-trivial to build mismatched voice-face pairs in deepfake datasets as opposed to generic datasets wherein identities are labeled. For improved adaptation to deepfake data, we pose a novel Real-Fake Disentangled loss, termed RFD, to effectively learn the matched and mismatched pair features on the basis of the real and fake samples.
% Besides, the computational resource requirement still falls within the practical remit, \textit{i.e.}, one can always synthesize just a few fake videos and fine-tune the model on them in minutes.

% Thereafter, this model could be smoothly transferred to detect forgery videos via fine-tuning on a small batch of deepfake data. Notably, the fine-tuning stage is efficient (only a few minutes and limited computational resource are required). Besides, the VFD alleviates the limitation of auxiliary data, which serves as another merit. 
% the deepfake data are agnostic to the training of VFD, thus avoiding any overfitting on deepfake features, and the model's generalizability over various deepfake datasets is thereby guaranteed. Besides, the VFD alleviates the limitation of auxiliary data or fine-grained deepfake artifacts, which serves as another merit. 

We conduct extensive experiments over three widely exploited deepfake datasets - DFDC, FakeAVCeleb~\cite{FakeAVCeleb}, and DeepfakeTIMIT~\cite{DFTIMIT}. The results demonstrate that our VFD achieves state-of-the-art performance, for instance, AUCs of 85.13\% and 86.11\% on DFDC and FakeAVCeleb, respectively. In addition, our method achieves remarkably competitive results with few deepfake data for fine-tuning as compared to some strong baselines. 

The main contributions of this work are three-fold:
\begin{itemize}
    \item 
    We address the deepfake detection from a voice-face matching view. To the best of our knowledge, we are the first to perform deepfake detection via the intrinsic correlation of facial and audio excluding any additional auxiliary data either more modalities or more visual features.
    \item 
    We devise an effective multi-modal matching framework to justify real and fake videos. On the basis of the matching view, we enhance the traditional contrastive loss to align the objective between generic and deepfake datasets.
    \item 
    Comprehensive quantitative and ablative experiments demonstrate that our method produces significant performance gains over a variety of SOTA competitors. Further experiments demonstrate that our model can still achieve a practical detection capability even with limited fine-tuning data.
    \end{itemize}

The rest of this paper is structured as follows. Section~\ref{sec:Related_work} briefly reviews the related literature. Method intuition and architecture are presented in Section~\ref{sec:Method}.
Section~\ref{Sec:exp} elaborates the experimental settings and results,  followed by the conclusion and discussion of this paper in Section~\ref{Sec:con}. 

%% file: 2_rel.tex
\section{Related Work}
\label{sec:Related_work}
% Our study aims to mainly contribute to determining the authenticity of the input faces on the basis of matching perspective. In the following, we study three literature streams related to our proposed model, \textit{i.e.}, deepfake~\cite{df1}, deepfake detection~\cite{dfd1}, and cross-modal biometric matching~\cite{NEURIPS2019_eb9fc349}. Moreover, we highlight the fact motivating us to explore this untapped view.
\subsection{Deepfake}
\label{Sec:Single-modal}
Benefiting from the continuous development of portrait synthesis, deepfake has recently emerged as a prevailing research problem. Existing algorithms either leverage the image only or the 3D information~\cite{Tip_cite_2} to edit videos. The image only methods synthesize fake faces for the target identities, which are then blended into the given video~\cite{DBLP:journals/tog/BitoukKDBN08}. For example, Li \textit{et~al.}~\cite{shifter} utilized cascaded AAD blocks to integrate identities and face attributes within multiple feature levels, and realistic human faces can then be generated. Wav2Lip~\cite{df6} synthesizes accurate speaking videos driven by speeches and the upper face. Different from these methods, Kim~\textit{et~al.}~\cite{DBLP:journals/tog/KimCTXTNPRZT18} applied 3DMM~\cite{3DMM2} to produce the portraits with controllable poses. HifiFace~\cite{DBLP:conf/ijcai/0002CZCTWLWHJ21} generates photo-realistic videos via the 3D shape-aware identity extractor. However, the existing deepfake approaches pay much attention to the face regions while the voice-face consistency is hard to be maintained. 

\subsection{Deepfake Detection}
\label{Sec:Cross-modal}
Deepfake detection is often cast as a binary (real or fake) classification task. Preliminary efforts often endeavor to detect the specific traces of manipulation~\cite{dfd1,F3Net}. Masi~\textit{et~al.}~\cite{dfd3} proposed a two-branch network to separately extract artifacts of color and frequency domains~\cite{DBLP:conf/cvpr/LiuLZCH0ZY21}. SSTNet~\cite{dfd4} detects edited faces through spatial, steganalysis, and temporal features. In contrary to these approaches utilizing the vision modality only, studies nowadays exploit the multi-modal information~\cite{DBLP:conf/ijcai/GuoNCJZB21} for deepfake detection~\cite{attm,MMDFD1, DBLP:conf/icassp/YangLL19a}. For instance, lip-syncing and dubbing models~\cite{DBLP:conf/eusipco/KorshunovM18} are employed to identify the audio-visual inconsistency from a speaker. Hou~\textit{et al.}~\cite{jointAV} predicted the probability of voices and faces being edited to judge the video credibility. Mittal \textit{et~al.}~\cite{emotion} extracted the emotional biases that video and audio jointly mention, based on which the detection objective can be achieved. Previous approaches have gained certain improvements on some datasets. Nonetheless, when transferring to unknown data, inferior performance is often confronted. 

To address this lack of generalization issue, several cross-dataset detection approaches are proposed. Li~\textit{et~al.}~\cite{x-ray} constructed auxiliary data from extracted blending regions in large-scale videos to enhance the robustness. Haliassos \textit{et~al.}~\cite{lipsdontlie} utilized a pre-trained lip-reading model to explore the irregularities in mouth movements, followed by delicate fine-tuning on the forgery data. Nonetheless, these methods always require auxiliary data and yield increased training overload. In this work, we propose to tackle deepfake detection from a novel matching view. Specifically, the matching between voices and faces is taken as the proxy for discriminating real and fake videos, since the voices and faces show a severe mismatch from deepfake algoritms. Our method exhibits promising generalization over various deepfake datasets, which requires only paired voices and faces for training.

\subsection{Cross-modal Biometric Matching}
Cross-modal biometric matching aims to retrieve the corresponding video for a given audio from multiple candidates, or vice versa~\cite{DBLP:conf/cvpr/NagraniAZ18}. Among the initial efforts, researchers extracted video and audio features via pre-trained models and then employed cross entropy~\cite{DBLP:conf/cvpr/NagraniAZ18} or cosine loss~\cite{DBLP:conf/mm/HoriguchiKN18, DBLP:journals/tip/GuoNCTZ22} to measure the matching degree. Later studies take into consideration the interactions among different modalities. For instance, Wen \textit{et~al.}~\cite{DBLP:conf/cvpr/WenXJ0HH21} devised a two-level loss, which leverages both local and global features on modality alignment. ADSM~\cite{DBLP:conf/mm/ChengLCWXZ20} adopts an adversarial matching network to extract the high-level semantical features. A specially designed discriminator is then employed to bridge the voice and face gap while maintaining semantic consistency. Moreover, Speech2Face~\cite{DBLP:conf/cvpr/OhDKMFRM19} applies a pre-trained face decoder network to reconstruct the face from speech clips. The methods in this category, indeed provide certain support that the voices and faces from the same person are strongly correlated.

%% file: 3_prem.tex
\section{Methodology}
\label{sec:Method}
In this section, we first present the evidence from two aspects for the intuition of our method, followed by a detailed introduction of our multi-modal matching model pre-trained on generic audio-visual data. We end this section with the fine-tuning and detection on downstream deepfake datasets. 

\begin{figure}[t]
    \centering
        \includegraphics[width=0.35\textwidth]{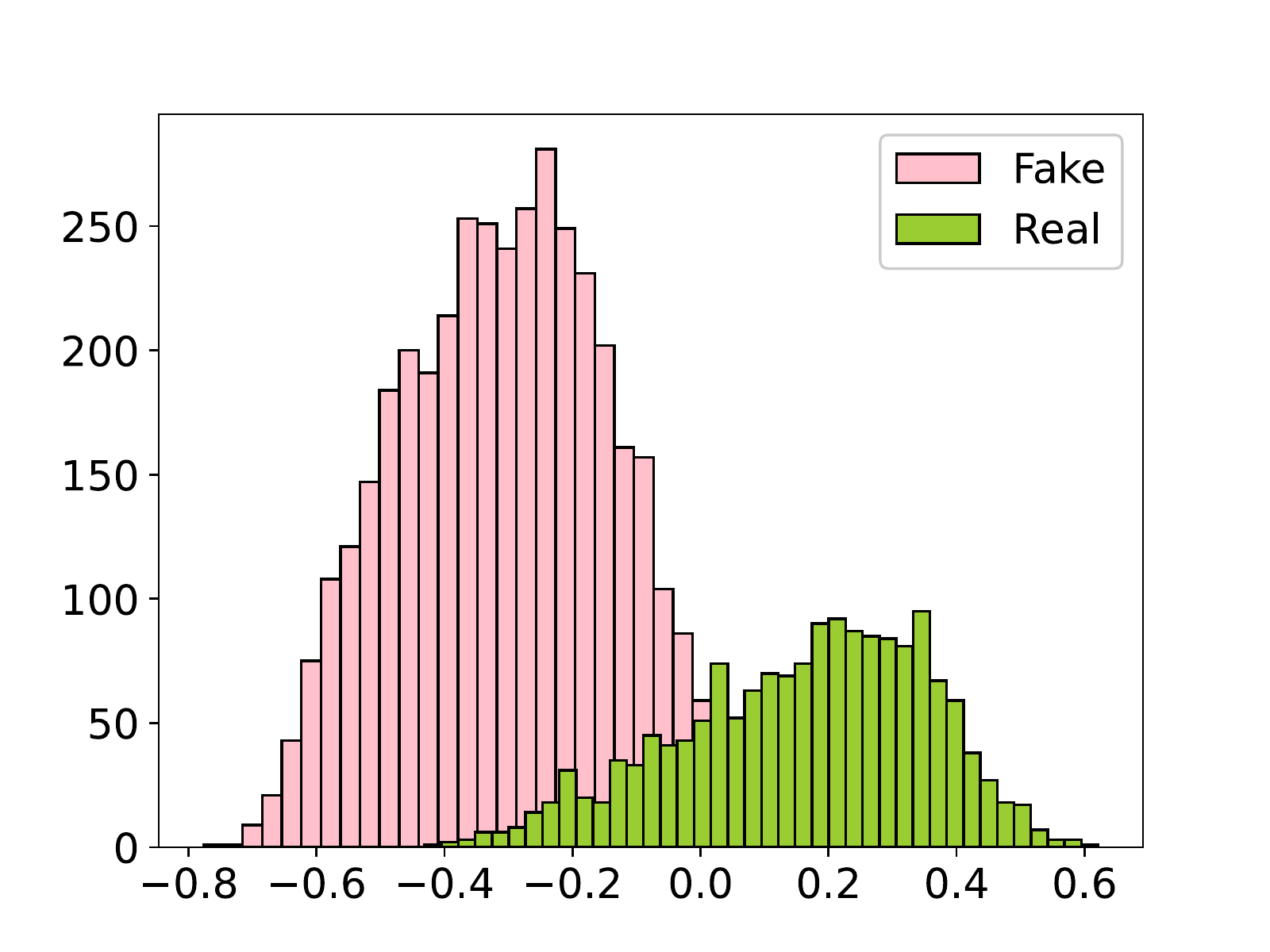}
    \caption{The similarity between voices and faces in fake (pink) and real (green) videos. The x-axis represents the similarity values, and the y-axis is the number of videos.}
    \label{new_distance} 
\end{figure}

\subsection{Method Intuition}
Research on human perception and neurology has shown that humans can outline pictures of a person's appearance based on the voices, or vice versa~\cite{KAMACHI20031709}. That is, voices and faces exhibit a high degree of homogeneity in the human brain and are associated closely with identities. In light of this, we contribute to exploring whether the matching between these two can serve as a proxy in detecting deepfake. To this end, we perform some probing tests, which mainly answer the following two questions:
\begin{itemize}
    \item 
    \textbf{Q1:} Are the voices and faces matched in deepfake videos?
    \item 
    \textbf{Q2:} Can the voices and faces be leveraged to discriminate different identities?
\end{itemize}

\subsubsection{Voice-Face Mismatching in Deepfake Data (Q1)}
\label{Section:unmatchDV}
We evaluate whether the voices and faces are matched in deepfake videos via measuring their similarity. In particular, we randomly sample 1,800 real and 4,000 corresponding fake videos from DFDC. Two plain transformer-based~\cite{VisionTrans,DBLP:conf/cvpr/NagraniAZ18} models are employed to extract the voice and face features. Thereafter, we calculate the corresponding cosine similarity between these two sets of features as follows,

\begin{equation}
\label{cos_simi}
\text { similarity }=\frac{\mathbf{v} \cdot \mathbf{f}}{\max \left(\left\|\mathbf{v}\right\|_{2} \cdot\left\|\mathbf{f}\right\|_{2}, \epsilon\right)},
\end{equation}
where $\mathbf{v}$ and $\mathbf{f}$ are voice and face features, respectively. $\epsilon$ is a small value to avoid division by zero.

We display the results in Figure~\ref{new_distance}. From this figure, we have the following observations: 1) The voice-face similarity in real videos is much larger than those in fake ones (the similarity split line is around $0.0$). That is, the voice and face from real videos match better in the feature space, while fake videos demonstrate serious mismatch evidence. And 2) a small fraction of real videos share similar values with fake ones, which is partially due to the presence of the ambient noise during recording.

\begin{figure*}[t]
    \centering
        \includegraphics[width=0.95\textwidth]{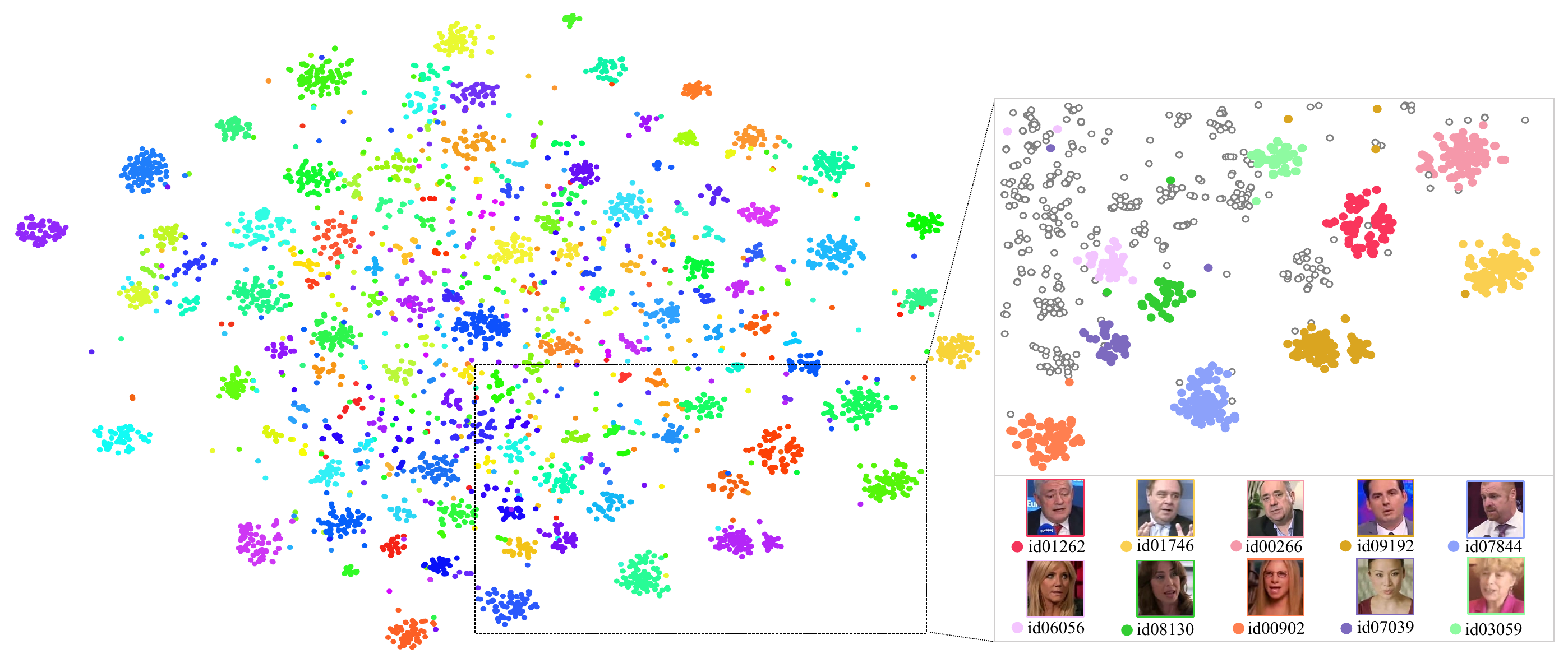}
    \caption{t-SNE~\cite{tsne} visualization of voice features from 163 identities. We zoom in the bottom right region and highlight ten identities with over 40 video instances. Note that that the color of each cluster is unique though some clusters may look visually same due to the excessive number of identities.}
    \label{audio_class_prob1} 
\end{figure*}

\subsubsection{Voice Discrimination over Identities (Q2)}
Faces encode essential cues for distinguishing different identities, which has been extensively proven by considerable studies~\cite{face_rec1_tip, face_rec2_tip, face_rec3_tip}. To testify whether the voices can achieve similar effects, we extract the voice features and cluster them in the following way. We employ the VGG-based~\cite{VGG_based} model as the backbone and the InfoNCE loss to learn the voice features, where the voice clips from the same identity are deemed as positive and from other identities as negative. In the second step, we randomly sample 5,120 voice clips from 163 people in the Voxceleb2 dataset and show the voice feature manifold embedding in Figure~\ref{audio_class_prob1}.

Figure~\ref{audio_class_prob1} tells that voices from the same identity tend to cluster together with apparent boundaries, demonstrating the discrimination capability of voices. For details, we zoom in the bottom right region and highlight ten identities whose associated videos are over 40. This detailed view illustrates that the learned features correlate tightly with group characteristics. For example, the female identities (\textit{id00656}, \textit{id08130}, \textit{id00902}, \textit{id07039}, and \textit{id03059}) distribute on the top left. Moreover, given the same gender, the voice features can also discrimina te ages. For example, \textit{id01262}, \textit{id01746}, and \textit{id00266} share similar ages (in their 60s), driving the voice features closer (top right region).

% \begin{table*}
% \centering
% \renewcommand\arraystretch{1.3}
% \caption{Notations used in this paper for the proposed VFD.}
% \begin{tabular}{ll}
% \toprule
% Notation              &Definition \\ \midrule
% Input data:                             & \\
% \quad$\mathcal{A}$, $\mathcal{V}$       & Instances of input audios and videos.\\ 
% \quad$\mathbf{C}$, $\mathbf{I}$         & Voice spectrogram $\mathbf{C}$, face frame $\mathbf{I}$. \\
% \quad$\mathbf{I}^+$, $\mathbf{I}^-$     & Manually selected positive and negative face frames for contrastive learning. \\
% \midrule
% Features:               &       \\
% \quad$\mathbf{I}_v$                     & Features after convolutional projection. \\
% \quad$\mathbf{c}$, $\mathbf{M}_I$       & Learnable vector $\mathbf{c}$, matrix $\mathbf{M}_I$ via concatenating $\mathbf{c}$ to $\mathbf{I}_v$. \\
% \quad$\mathbf{E}$, $\mathbf{M}_p$       & Position embedding matrix $\mathbf{E}$, matrix $\mathbf{M}_p$ via adding $\mathbf{E}$ to $\mathbf{M}_I$. \\
% \quad$\mathbf{M}_q$, $\mathbf{M}_k$, $\mathbf{M}_k$  & Projection matrices in the feature extractors.\\
% \quad$\mathbf{K}_I$, $\mathbf{f}_f$     & Final matrix $\mathbf{K}_I$ by mapping projection matrices, feature vector $\mathbf{f}_f$ \\
% \bottomrule 
% \end{tabular}
% \label{ablation}
% \end{table*}

\subsection{VFD for Pre-training}
\begin{figure*}[t]
    \centering
        \includegraphics[width=0.95\textwidth]{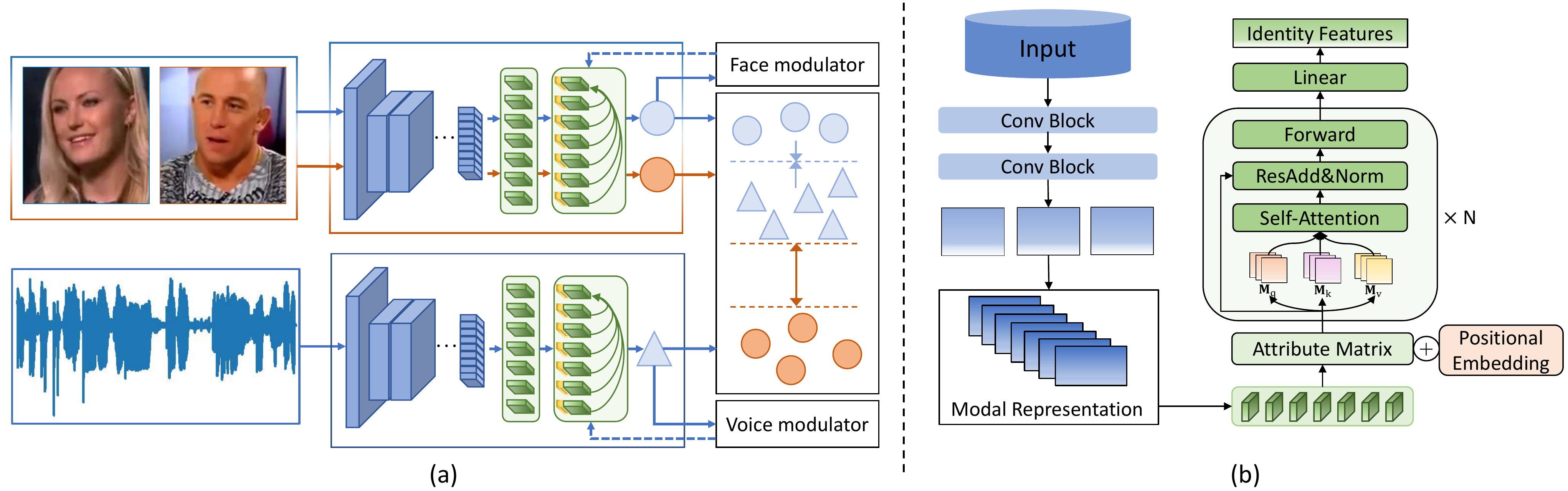}
    \caption{Overall architecture of (a) Voice-face matching network, (b) voice/face feature extractor.}
    \label{MMN} 
\end{figure*}

Based on the above findings that a single person's voices and faces have certain homogeneity, in this paper, we propose to detect deepfake videos by judging the matching degree of these two. To achieve this goal, we design a Voice-Face matching Detection (VFD) method. As shown in Figure~\ref{MMN}a, our VFD is embodied with a dual-stream network, wherein the voices and faces are separately processed. Two modulators, \textit{i.e.}, face and voice modulator, are employed to guide the extractors to focus on the identity-related features, followed by a matching function, \textit{e.g.}, InfoNCE loss~\cite{InfoNCE}, to determine the matching degree. 

We first train the model on a generic audio-visual dataset. Our motivation for implementing pre-training stems from two aspects: 1) Learning cross modality homogeneity has long been recognized non-trivial, which demands large quantity of multi-modal paired data~\cite{DBLP:conf/cvpr/WenXJ0HH21, DBLP:journals/tip/YaoZWLYL22}. And 2) different deepfake datasets focus on distinctive forgery angles and the volume of one dataset is often less sufficient to train a generalizable matching model~\cite{dfdc,DBLP:conf/interspeech/NagraniCZ17,FakeAVCeleb}. Therefore, directly training VFD leads to sub-optimal performance, as revealed in Section~\ref{abl}. We instead, train our VFD on a generic dataset. In what follows, we first present the data preprocessing protocol. After that, we elaborate on the overall architecture of our VFD and its corresponding pre-training strategy sequentially.

\subsubsection{Data Preprocessing}
\label{Secton:Data_prepro}
For each input audio $\mathcal{A}^{}_{}$, we extract a three-second voice clip with a 16kHz sample rate, which will be represented as a spectrogram $\mathbf{C} \in \mathbb{R}^{\mathrm{L}_1 \times \mathrm{H}_1 \times \mathrm{W}_1}$, where $\mathrm{L}_1$ denotes the channel number, $\mathrm{H}_1$ and $\mathrm{W}_1$ are the height and width of the spectrograms, respectively. On the other hand, for the video input $\mathcal{V}^{}_{}$, we randomly leverage one random face frame $\mathbf{I} \in \mathbb{R}^{\mathrm{L}_2 \times \mathrm{H}_2 \times \mathrm{W}_2}$ to represent it as each video involves one identity only~\cite{voxceleb2,dfdc,FakeAVCeleb}.

The key to performing contrastive learning is to construct efficacious positive and negative samples. In the pre-training, we utilize the voice as the anchor and build \textit{$\textless$positive, negative$\textgreater$} pairs from identity faces\footnote{It is also feasible to choose faces as anchors, which will serve as a possible extension of this work.}. Specifically, for a given voice clip $\mathbf{C}$, it is straightforward to sample the faces from the same identity as its positive $\mathbf{I}^{+}$. As to the negative $\mathbf{I}^{-}$ sampling, we simply adopt the faces from other identities as the counterpart while leaving cumbersome hard negative mining as future work. After this, the voice and face features (both positive and negative) are inputted to two independent feature extractors, which is detailed as follows.

\subsubsection{Feature Extraction}
\label{pretraining}
Extracting both voice and face features associated with identity is of vital importance for a multi-modal matching approach. To this end, an elaborated multi-modal feature extractor is presented, and the overall architecture is shown in Figure~\ref{MMN}b. In the next, we take the processing for the face input $\mathbf{I}$ as an example, and the voice feature extractor is designed in a similar fashion. 

% inherits from a well-developed VGG-Face model~\cite{vgg-face}
We firstly utilize a deep forward convolutional projection to cut the faces into patches and extract the initial face features,
\begin{equation}
\mathbf{I}_{v}(i, j, m)=\sum_{u, v=0}^{\mathrm{N}} \sum_{t=0}^{\mathrm{F}} \operatorname{h}(u, v, t) \otimes \mathbf{I}(i-u, j-v, m-t),
\end{equation}
where $\mathrm{N}$ is the image size (the height and width are identical in face images), $\mathrm{F}$ denotes the output channel number, and $\operatorname{h}(\cdot)$ represents the convolutional kernel. In this way, we obtain the F-channel feature maps $\mathbf{I}_v \in \mathbb{R}^{\mathrm{F} \times \mathrm{D}}$, representing the modality features.

It is non-trivial to learn the overall identity features of voices or faces with the above module, as only local information is aggregated by this fashion while the global context is overlooked. Therefore, we develop another novel transformer-like~\cite{VisionTrans} module to learn the identity features based on the initial multi-modal representations, wherein the self-attention mechanism is employed to extract non-local joint features. To this end, we first build a new feature matrix $\mathbf{M}_l$ via concatenating a learnable vector $\mathbf{c} \in \mathbb{R}^\mathrm{D}$ with $\mathbf{I}_v$ to gather identity-related features from the modal representations. Subsequently, the newly constructed map $\mathbf{M}_l$ will be element-wisely added by a positional encoding $\mathbf{E}$ for retaining the positional information of the image feature maps,
\begin{equation}
\mathbf{M}_p= \mathbf{M}_l \oplus \mathbf{E},
\end{equation}
where $\mathbf{M}_p, \mathbf{M}_l$, and $\mathbf{E} \in \mathbb{R}^{\mathrm{F}^{'} \times \mathrm{D}}$, $\mathrm{F}^{'} = \mathrm{F} + 1$. 

And then, we map $\mathbf{M}_p$ into three matrices,
\begin{equation}
[\mathbf{M}_q, \mathbf{M}_k, \mathbf{M}_v] = [\mathbf{W}_q, \mathbf{W}_k, \mathbf{W}_v] \otimes \mathbf{M}_p,
\end{equation}
where these three matrices share a same size of ${\mathrm{F}^{'} \times \mathrm{D}}$. We then utilize the self-attention mechanism to perform non-local learning, 
\begin{equation}
\operatorname{g}(\mathbf{M}_q, \mathbf{M}_k, \mathbf{M}_v)=\operatorname{softmax}\left(\frac{\mathbf{M}_q \mathbf{M}_k^{\mathrm{T}}}{\sqrt{\mathrm{D}}}\right) \mathbf{M}_v.
\end{equation}

We perform this block several times and finally map the outputs of $\operatorname{g}(\cdot, \cdot, \cdot)$ into the $\mathbf{K}_f \in \mathbb{R}^{\mathrm{F}^{'} \times \mathrm{V}}$ with a linear transformation. In this calculation, the vector $\mathbf{c}$ has aggregated the global identity-related information via the self-attention mechanism~\cite{VisionTrans}. We thus detach $\mathbf{c}$ from $\mathbf{K}_f$ and reformulate this vector as $\mathbf{f}_{f} \in \mathbb{R}^{V}$ to represent the features of the whole face frame. In a similar fashion, the voice branch will transform voice input to $\mathbf{K}_v$ with the same shape as $\mathbf{K}_f$, and a feature vector $\mathbf{f}_{v}$ would be selected as the representative. Note that due to the modality gap between voices and faces~\cite{DBLP:conf/cvpr/NagraniAZ18}, we do not share parameters between these two extractors but instead expect the InfoNCE loss (see following) to enhance the homogeneity modeling. 

\subsubsection{Pre-training Protocol}
We employ three loss functions to update the parameters of VFD. The first two are naive cross-entropy loss, and the last one is the InfoNCE loss~\cite{InfoNCE} for judging matchness.

The cross-entropy loss is applied to both the voice and face modulators for binding identity information to the feature vectors. Specifically, we utilize the identity in the given video as the label for feature extraction,
\begin{equation}
\mathcal{L}_{cls}=-\log \frac{\exp \left(\mathbf{W}_y\mathbf{f}\right)}{\sum_{i=1}^{\mathrm{C}} \exp \left(\mathbf{W}_i\mathbf{f}\right)},\\
\end{equation}
where $\mathrm{C}$ is the number of identities, $\mathbf{W}$ is a weighted matrix to calculate the feature vector $\mathbf{f}_{v}$ or $\mathbf{f}_{f}$ to the propensity of a particular identity, and $y$ is the label. In this manner, the voice and face extractors will be empowered to extract identity-associated features from the input modalities. Thereafter, the InfoNCE loss~\cite{InfoNCE} is applied to achieve the homogeneity as follows:
\begin{equation}
\label{tri_loss}
\mathcal{L}_{I} \! = \! \mathbb{E}_{\mathbf{f}_f \sim p, \mathbf{f}_f^{+} \sim p^{+}}\left[-\operatorname{log} \frac{e^{\operatorname{d}(\mathbf{f}_v, \mathbf{f}_{f}^{+})/\tau}}{e^{\operatorname{d}(\mathbf{f}_v, \mathbf{f}_f^{+})} + \sum_{i=1}^{\mathrm{U}} e^{\operatorname{d}(\mathbf{f}_v, \mathbf{f}_{fi}^{-})}}\right],
\end{equation}
where $\mathbf{f}_f^+$ corresponds to a positive instance feature extracted from the matched face image $\mathbf{I}^+$; $\mathbf{f}_{fi}^{-}$ is the negative instance featuer as described in \ref{Secton:Data_prepro}; $\mathrm{U}$ is the number of negativa samples; $\tau$ represents a hyperparameter temperature, and $\operatorname{d}(\cdot,\cdot)$ is defined as the cosine similarity in the joint space between voices and faces as shown in Equation~\ref{cos_simi}.

By means of calculating the mutual information between $\operatorname{d}(\mathbf{f}_v, \mathbf{f}_f^+)$ and $\operatorname{d}(\mathbf{f}_v, \mathbf{f}_{f}^-)$, VFD pulls the matched voice clip $\mathbf{f}_v$ and the positive faces $\mathbf{f}_f^+$ closer while pushing the unmatched negative faces $\mathbf{f}_{f}^-$ further. The $\tau$ contributes to make the positive and negative instances more separable.

\subsection{VFD for Fine-tuning}
\label{finetuning}
% Based on the aforementioned architecture, our VFD is pre-trained on a generic audio-visual dataset with the potential to model identity-related features homogeneously. Thereafter, the most intuitive way is to apply this model to downstream deepfake data directly. However, this setting can lead to sub-optimal performance since it neglects that the mismatched voice-face pairs in the pre-training stage are manually constructed rather than deepfake algorithms (Section~\ref{abl} will discuss this setting). 

Benefiting from the large-scale pre-training, we then leverage the pre-trained model for downstream specific deepfake detection. Inspired by prior work~\cite{x-ray}, we advocate one that quickly adapts in a fine-tuning fashion to improve the model generalization. In particular, We transfer the pre-trained VFD model to deepfake videos where voices and faces might be from some mismatched identities. Notably, the pre-training has made the model quite sensitive to mismatched voices and faces. Therefore, applying part of the deepfake dataset in the fine-tuning stage is feasible to quickly adapt the model to deepfake data, wherein significant computational overhead is thereby circumvented.

During fine-tuning, we jointly optimize the voice and face extractor such that the voice features and its matched face ones emit the highest similarity score among others. As the deepfake dataset contains both real and fake voice-face pairs, it thus releases the pain of negative sampling in Equation~\ref{tri_loss} since the fake pairs are in fact mismatched. In other words, instead of manual selection of mismatched faces as negative samples, we could refer to the real and fake videos as positive and negative samples, respectively, to fine-tune the pre-trained model in a contrastive learning fashion. However, this setting throws up a new issue, \textit{i.e.}, \textbf{it is challenging to confirm that the anchor voice $\mathbf{f}_v$ from real video and negative face $\mathbf{f}_{f}^{-}$ from fake ones are mismatched since the voices and faces are extensively edited, and the traditional InfoNCE loss may thereby confused.} Therefore, to further align the objective between pre-training and fine-tuning, we upgraded classic contrastive loss to a simple real-fake contrastive loss function (RFC). Specifically, we first compute the similarity of voices and faces from positive and negative instances:
\begin{equation}
\label{logits}
\left\{
\begin{aligned}
&v_{pos} = \operatorname{d}(\mathbf{f}_v^+, \mathbf{f}_f^+), \\
&v_{neg} = \operatorname{d}(\mathbf{f}_v^-, \mathbf{f}_f^-),
\end{aligned}
\right.
\end{equation}
wherein we calculated the similarity via the the priori labels - real or fake - in the deepfake dataset rather than fixed anchors. We thereby model the homogeneity with a simple cross-entropy manner:
\begin{equation}
\label{RFD}
\begin{aligned}
\mathcal{L}_{RFC}=
-\sum \mathbf{t} \log \left(\operatorname{softmax}([v_{pos}, v_{neg_{i}}|_{i=1}^{\mathrm{Q}}])\right),
\end{aligned}
\end{equation}
where the $\mathrm{Q}$ is the number of negative samples; label vector $\mathbf{t} \in \mathbb{R}^{\mathrm{Q}'}, \mathrm{Q}'=\mathrm{Q}+1$ is artificially generated tags wherein $\mathbf{t}_i = 1$ for positive similarity $v_{pos}$, as opposed to $\mathbf{t}_i = 0$ for negative ones. RFC can effectively pull up the similarity of positive pairs relative to the negative ones, \textit{i.e.}, the voices and faces in the real video will match better, while the fake ones will exhibit a high degree of inconsistency. 
In a nutshell, RFC is an upgraded version of vanilla methods such as InfoNCE on the deepfake dataset, which estimates the mismatch between the fake voices and faces and avoids the consumption of manually selecting negative examples.

\subsection{VFD for Deepfake Detection}
On the basis of the aforementioned protocols, our VFD is pre-trained and then fine-tuned on generic audio-visual dataset and deepfake dataset, respectively. For deepfake detection, we extract the face images $\mathbf{I}_d$ and voice clips $\mathbf{C}_d$, and apply VFD to determine the matchness:
\begin{equation}
\label{matching}
\text{Matching}=\left\{
\begin{aligned}
&True  , & \text{VFD}(\mathbf{C}_d, \mathbf{I}_d) \geq \lambda, \\
&False  , & \text{VFD}(\mathbf{C}_d, \mathbf{I}_d) < \lambda,
\end{aligned}
\right.
\end{equation}
where the output of VFD($\cdot$, $\cdot$) is the similarity computed via Equation~\ref{cos_simi} in the joint latent space; and $\lambda$ is selected from the validation set. According to Equation \ref{matching}, the matching result of the $\mathbf{I}_d$ and $\mathbf{C}_d$ being $True$ denotes a matched voice-face pair, namely, the input video is real. On the contrary, the $False$ result corresponds to fake videos edited by deepfake algorithms.

%% file: 4_exp.tex
\section{Experiment}
\label{Sec:exp}
\subsection{Dataset}
We utilized the Voxceleb2 dataset for pre-training of our VFD model. Specifically, Voxceleb2 is a generic audio-visual dataset collected from YouTube videos, containing over 1 million utterances from 6,112 celebrities. We split them into training, validation, and testing sets with a ratio of 8:1:1. Thereafter, three deepfake datasets - DeepfakeTIMIT, DFDC, and FakeAVCeleb, are employed to fine-tune and evaluate the effectiveness of VFD.
\begin{itemize}
    \item DeepfakeTIMIT~\cite{DFTIMIT} contains two subsets of fake videos, namely, lower quality (LQ) via a 64 $\times$ 64 input/output size model, and higher quality (HQ) with 128 $\times$ 128 one. Each subset involves 16 similar looking pairs of subjects, wherein each subject has 10 face-swapping videos. We selected the original videos from VidTIMIT~\cite{VIDTIMIT} and all fake ones from DeepfakeTIMIT for our fine-tuning and testing. For each subset, we divided the videos into fine-tuning and testing sets according to the ratio of 4:1.
    \item DFDC~\cite{dfdc} contains 23,654 real videos recorded from 960 identities and 104,500 fake videos. Since some videos in this dataset are mixed with the camera holders’ voices, we manually filtered 6,089 real and 32,245 fake videos as the fine-tuning set, while 1,700 real videos with corresponding 5,810 fake videos are considered the testing set. 
    \item As to the FakeAVCeleb~\cite{FakeAVCeleb}, a total of 500 real videos and 19,500 fake videos are included, of which we used 391 real videos and the corresponding 16,869 fake videos as the fine-tuning set and the remaining part as the testing set. It is worth noting that the pre-training, fine-tuning, and testing sets in the datasets mentioned above do not contain any duplicated identities.
\end{itemize}

% 1700 + 859
% 5810 + 3783

% Since some videos in this dataset are mixed with the camera holders' voices, we manually filtered  as the testing set. 

\subsection{Implement Details}
% ~\cite{DBLP:journals/tip/GuoNCTZ22}
We implemented our model with the Pytorch toolkit. And the AdamW optimizer~\cite{AdamW} is adopted with base learning rate of $1\times10^{-4}$ and $5\times10^{-6}$ for parameter updating of pre-training and fine-tuning, respectively. The learning rate decays gradually following the training process with the weight decay $0.2$. All model parameters are initialized using a random normal distribution with a mean of 0 and a standard deviation of 0.02. The model is trained with a mini-batch size of 128 on 4 Tesla V100 GPUs. The threshold $\epsilon$ used in Equation~\ref{cos_simi} is $1\times10^{-8}$, the $\tau$ employed in Equation~\ref{tri_loss} is set at $0.1$, and the $\lambda$ in Equation~\ref{matching} is $-0.1$ as selected in the validation set of Voxceleb2. We used 12 transformer blocks with 12-head self-attention in identity feature extractors, \textit{i.e.}, the $\mathrm{N}=12$ in Figure~\ref{MMN}b. The input face images are resized to $224 \times 224$, while the voice clips are represented as $512 \times 300$ spectrograms since each second of voice clips is divided into 100 small windows in the sliding window manner.

\begin{table*}
\centering
\renewcommand\arraystretch{1.3}
\caption{Performance (\%) of VFD and baselines on DFDC, FAkeAVCeleb, and DeepfakeTIMIT. For comparison purposes, we reported the AUC on two subsets of DeepfakeTIMIT, namely LQ and HQ, follow previous work~\cite{FWA}. ${\ddag}$: the model is reproduced by ourselves; -: the authors did not report this metric on this dataset; The multi-modal models and their uni-modal versions are marked \textbf{pink}, while the fine-tuning ones and their fine-tuning-free counterparts are marked \textbf{blue}.}
\begin{tabular}{lccccccccccccc}
\toprule
\multicolumn{1}{c}{\multirow{2}{*}{Model}}      & \multicolumn{2}{c}{Modality}      &\multicolumn{1}{c}{\multirow{2}{*}{Auxiliary}}   & \multicolumn{2}{c}{DFDC}                              & \multicolumn{2}{c}{FakeAVCeleb} & \multicolumn{2}{c}{DeepfakeTIMIT} \\ \cmidrule(lr){2-3} \cmidrule(lr){5-6} \cmidrule(lr){7-8} \cmidrule(lr){9-10}
\multicolumn{1}{c}{}                             & Visual           & Audio                 &\multicolumn{1}{c}{}             & ACC               & AUC             & ACC      & AUC    & LQ        & HQ    \\ \midrule
Meso-4\cite{MesoNet}~$^{\ddag}$                                    & \checkmark       & $\times$                  &$\times$                     & {49.23}           & 52.92           & {43.65}      & 49.17  &62.10  & 55.25\\
\rowcolor[HTML]{FFE6E6} MesoInception-4\cite{MesoNet}~$^{\ddag}$   & \checkmark       & $\times$                  &$\times$                      & {56.37}           & 60.56         & {72.22}      & 75.82   &78.45  & 60.70 \\
\rowcolor[HTML]{FFE6E6}EfficientNet\cite{efficient}                & \checkmark       & $\times$                  &$\times$                     & {-}               & {-}             & 81.03       &  -      &-  & -\\ 
\rowcolor[HTML]{FFE6E6}VGG16\cite{VGG_based}                       & \checkmark       & $\times$                  &$\times$                     & {-}               & {-}             & 59.64       &  -   &-  & - \\
Capsule\cite{capsule}~$^{\ddag}$                                   & \checkmark       & $\times$                  &$\times$                     & {57.65}           & 61.20           & {73.27}      & 76.19   &84.58  & 81.69\\
\rowcolor[HTML]{DAEAF1}Xception\cite{dfd1}~$^{\ddag}$              & \checkmark       & $\times$                  &$\times$                     & {73.09}           & 75.52           & {71.67}      & 76.19  &97.90  &95.48\\
\rowcolor[HTML]{DAEAF1}F$^3$-Net\cite{F3Net}~$^{\ddag}$            & \checkmark       & $\times$                  &$\times$                     & {74.16}           & 75.40           & 81.08      & 84.54  &98.39  &94.60\\
\rowcolor[HTML]{DAEAF1}ViT\cite{VisionTrans}~$^{\ddag}$            & \checkmark       & $\times$                  &$\times$                     & {74.97}           & 76.05           &{74.35}      & 80.49 &99.59  &98.84\\ \midrule
BA-TFD\cite{BA-TFD}                                                & \checkmark       & \checkmark                &$\times$                    & {-}               & 84.60           & {-}         &  -    &-  & -\\
Emotional Forensics\cite{emotion}                                  & \checkmark       & \checkmark                &Emotion                      & {-}               & 84.40           & {-}         &  -    &96.30  & 94.90 \\
\rowcolor[HTML]{FFE6E6}MesoInception\_MM\cite{mm_base}             & \checkmark       & \checkmark                &Ensemble                    & {-}               & {-}             & 72.87       &  -   &-  & -\\ 
\rowcolor[HTML]{FFE6E6}EfficientNet\_MM\cite{mm_base}              & \checkmark       & \checkmark                &Ensemble                     & {-}               & {-}            & 63.18       &  -   &-  & -\\
\rowcolor[HTML]{FFE6E6}VGG16\_MM\cite{mm_base}                     & \checkmark       & \checkmark                &Ensemble                      & {-}               & {-}           & 78.04       &  -   &-  & -\\  \midrule
VA-MLP\cite{VA-MLP}                                                & \checkmark       & $\times$                  &Landmark                     & -                 & 61.90           & -           & 67.00  &61.40  & 62.10\\
VA-LogReg\cite{VA-MLP}                                             & \checkmark       & $\times$                  &Landmark                     & -                 & 66.20           & -           & 67.90  &77.00  & 77.30\\
FWA\cite{FWA}                                                      & \checkmark       & $\times$                  &Landmark                     & -                 & 72.70           & -           & -  &99.90  & 93.20\\
DSP-FWA\cite{FWA}                                                  & \checkmark       & $\times$                  &Landmark                     & -                 & 75.50           & -           & -  &99.90  & 99.70\\
Headpose\cite{DBLP:conf/icassp/YangLL19a}                          & \checkmark       & $\times$                 &Landmark                     & {-}           & 55.90               &{-}          & 49.00 &55.10  & 53.20\\ \midrule
\rowcolor[HTML]{DAEAF1}Xception\_F~$^{\ddag}$            & \checkmark     & $\times$            &Fine-tune           & {67.84}          & 71.13            &  70.53  & 71.20  &95.17  & 93.69 \\
\rowcolor[HTML]{DAEAF1}F$^3$-Net\_F~$^{\ddag}$           & \checkmark     & $\times$            &Fine-tune           & 71.29          & 72.76            &  75.13  & 78.46  &98.25  & 96.66 \\
\rowcolor[HTML]{DAEAF1}ViT\_F~$^{\ddag}$                 & \checkmark     & $\times$                          &Fine-tune            &65.04 &71.16 &70.05 & 74.30   &99.00  & 98.17 \\
Face~X-ray (HRNet-18-BI100K)\cite{x-ray}~$^{\ddag}$      & \checkmark  & $\times$                             &Blending+Fine-tune     & {43.42}      & 59.36           & {72.88}     &  73.52  &96.95  & 94.47\\ 
Face~X-ray (HRNet-18-BI500K)\cite{x-ray}~$^{\ddag}$      & \checkmark  & $\times$                             &Blending+Fine-tune     & {44.80}      & 58.98           & {75.65}     &  77.94  &98.61  & 97.54\\ 
Face~X-ray (HRNet-32-BI100K)\cite{x-ray}~$^{\ddag}$      & \checkmark  & $\times$                             &Blending+Fine-tune     & {46.49}      & 61.57           & {76.75}     &  79.72  &99.72  & 98.53\\ 
LipForensics\cite{lipsdontlie}                           & \checkmark     & $\times$                          &Landmark+Fine-tune           & {-}          & 73.50       &  -  & -   &  -  & -\\
\midrule
VFD                                        & \checkmark       & \checkmark                      &Fine-tune        & {\textbf{80.96}}  & \textbf{85.13}   & \textbf{{81.52}}       & \textbf{86.11} & \textbf{99.95}  & \textbf{99.82}\\ 
\bottomrule
\end{tabular}
\label{performance}
\end{table*}

\subsection{Compared Baselines and Evaluation Metrics}
We compared our model with multiple state-of-the-art baselines regarding the metrics of ACC and AUC scores. They can be roughly divided into four groups:
\begin{itemize}
    \item 
    \textbf{Single modality} methods detect visual artifacts based on the vision modality only: 
    1) MesoInception-4 and 2) Meso-4~\cite{MesoNet},
    3) Capsule~\cite{capsule},
    4) F$^3$-Net~\cite{F3Net},
    5) Xception~\cite{dfd1},
    6) ViT~\cite{VisionTrans},
    7) EfficientNet~\cite{efficient}, and 
    8) VGG16~\cite{VGG_based}.
    \item
    \textbf{Visual-auditory detection} models: 9) BA-TFD~\cite{BA-TFD}, and 10) Emotional Forensics~\cite{emotion}. In addition, the uni-modal baselines 
    11) MesoInception-4\_MM, 
    12) EfficientNet\_MM, 
    and 13) VGG16\_MM have been reformulated as multi-modal versions via ensembling different classifiers~\cite{mm_base}. 
    \item
    \textbf{Models using auxiliary data}, \textit{e.g.}, the facial landmarks, guided manipulation traces detection:
    14) VA-MLP and 15) VA-LogReg~\cite{VA-MLP}, 
    16) FWA and DSP-FWA~\cite{FWA}, 
    17) Headpose~\cite{DBLP:conf/icassp/YangLL19a}.
    \item
    \textbf{Fine-tuning strategy} based approaches, \textit{i.e.}, 
    18) Face X-ray\cite{x-ray}, and 
    19) LipForensics~\cite{lipsdontlie}. Besides, we reproduced the traditional three uni-modal approaches, namely, 20) Xception\_F, 21) F$^3$-Net\_F, and 22) ViT\_F, with fine-tuning method, wherein the first two are pre-trained on FF++ dataset, while the ViT\_F possesses the identical training strategy with VFD, \textit{i.e.}, pre-training on Voxceleb2 dataset and fine-tuning on DFDC and FakeAVCeleb.
\end{itemize}
% eight baselines detect visual artifacts in a single modality fashion, namely, 
% MesoInception-4 and Meso-4~\cite{MesoNet},
% Capsule~\cite{capsule},
% F$^3$-Net~\cite{F3Net},
% Xception~\cite{dfd1},
% ViT~\cite{VisionTrans},
% EfficientNet~\cite{efficient},
% and VGG16~\cite{VGG_based}. Moreover, VA-MLP and VA-LogReg~\cite{VA-MLP},
% FWA and DSP-FWA~\cite{FWA}, Headpose~\cite{DBLP:conf/icassp/YangLL19a} are designed to detect manipulation traces with auxiliary data, for instance, the facial landmarks. Furthermore, BA-TFD~\cite{BA-TFD}, and Emotional Forensics~\cite{emotion} are proposed to identify fake videos with multi-modal information such as visual-audio cues. In addition, the uni-modal baselines MesoInception-4, EfficientNet, and VGG16 have been reformulated as multi-modal versions via ensembling different classifiers~\cite{mm_base}. 
% And the remaining two baselines, \textit{i.e.}, Face X-ray~\cite{x-ray} and LipForensics~\cite{lipsdontlie}, are models on the basis of fine-tuning strategy. Besides, we reproduced the traditional three uni-modal approaches, namely, Xception\_F, F$^3$-Net\_F, and ViT\_F, with fine-tuning method, wherein the first two are pre-trained on FF++ dataset, while the ViT\_F possesses the identical training strategy with VFD, \textit{i.e.}, pre-training on Voxceleb2 dataset and fine-tuning on DFDC and FakeAVCeleb.

\subsection{Performance Comparison}
\label{Per}
The results of baselines and our method on DFDC and FakeAVCeleb are demonstrated in Table~\ref{performance}. It can be observed that VFD achieves state-of-the-art performance on all datasets, expressing the effectiveness of our method as well as the validity of tackling deepfake detection using the voice-face matching view. Among these three datasets, DFDC and FakeAVCeleb are more challenging due to the inclusion of considerable videos and the plural forgery algorithms. Our VFD achieves around 85\% AUCs, while the baselines even show accuracy below 50\%, \textit{e.g.}, 49.23\% and 43.65\% of meso4. In contrast, the performance of baselines on DeepfakeTIMIT is more satisfying, \textit{i.e.}, most baselines are over 90\% AUC. This relates to the fact that DeepfakeTIMIT contains standard faces without occlusion or rotation from a naive deepfake approach. Therefore, the models can easily capture minor visual and auditory artifacts. In what follows, we will analyze the experimental results in different groups.

\subsubsection{Comparison on modality}
We can observe that some single-modality models perform unfavorably, with Capsule's accuracy less than 60\% and Meso-4's around 50\%. One possible reason for this is that the single-modality approaches rely heavily on the artifacts extraction capability of the backbones. Hence some strong models are prone to fail on the latest data with more realistic visual artifacts. In the multi-modal model zoo, the audio or emotion-based multi-modal methods, namely BA-TFD and Emotional Forensics, can significantly outperform the single-modality ones, demonstrating the promising potential of multi-modal cues. However, the ensembling models show both enhancements and decreases over their uni-modal counterparts (pink base color). We attributed this phenomenon to the fact that the ensembling model is designed to fuse the prediction of the two plain classifiers without considering the relations among modalities, which would lead to unstable performance~\cite{mm_base}. Therefore, the multi-modal detection models and training strategies must be carefully designed to exploit the complementary properties of multiple modalities.

\begin{figure}
    \centering
    \includegraphics[width=0.47\textwidth]{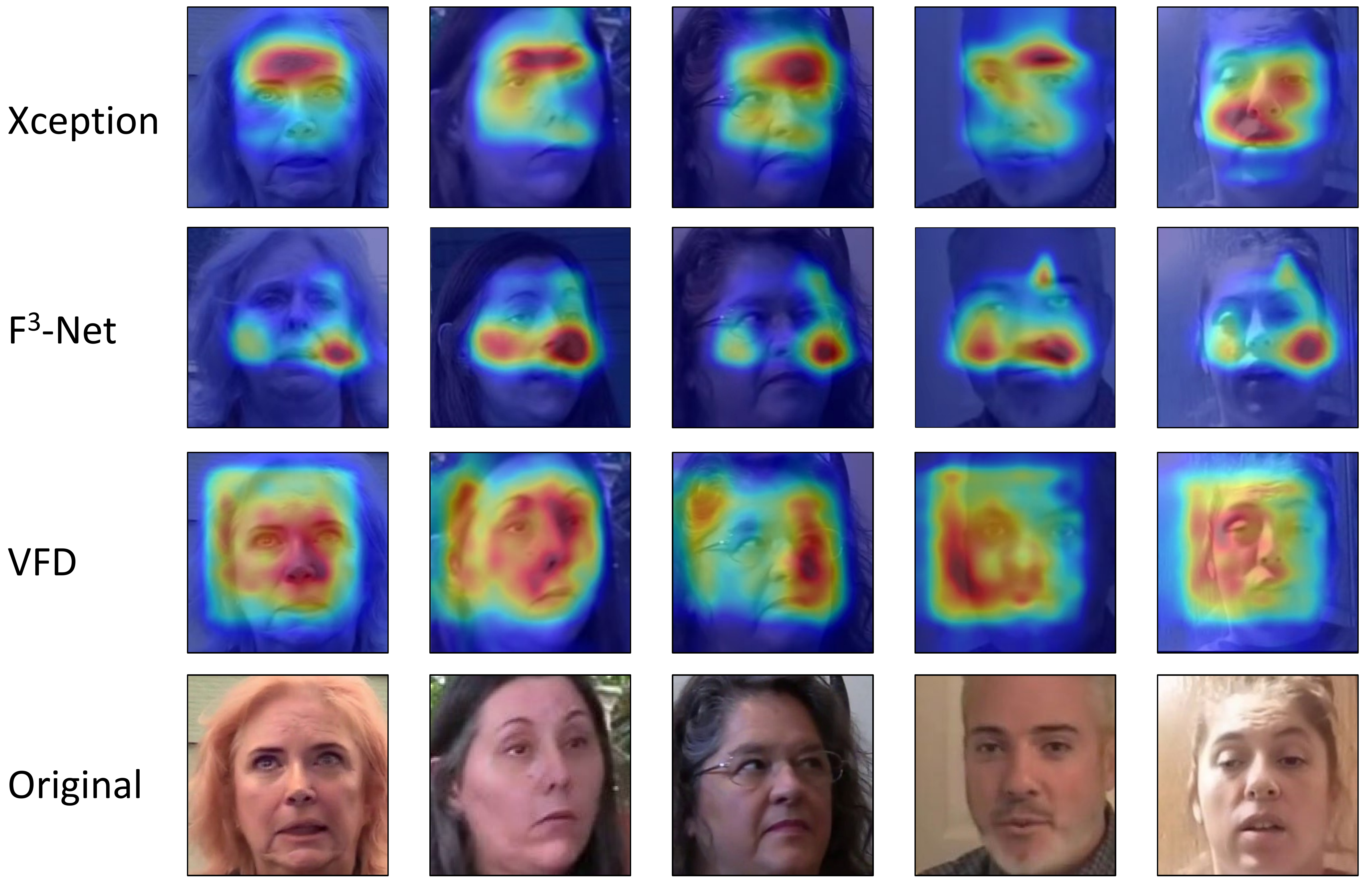}
\caption{Heatmaps produced by baselines and our method. The top two rows are from Xception and F$^3$-Net, followed by the third row of heatmaps yielded by our VFD and the original face images.}
\label{same_DFDC} 
\end{figure}

\subsubsection{Comparison on auxiliary data}
The models that employ auxiliary data, such as landmarks, perform comparably to traditional uni-modal methods. For example, DSP-FWA gains AUC over 75\% and 99\% on DFDC and DeepfakeTIMIT, which is on par with ViT, while Headpose only outperforms Meso-4. One possible reason is that the landmarks are representations of facial movements, while some powerful visual models may be sufficient to capture such features and thus achieve proximate performance.

\subsubsection{Comparison on fine-tuning strategy}
Out of expectation, the fine-tuning on traditional approaches is harmful (blue base color). For instance, Xception decreases 4.99\% on FakeAVCeleb, while F$^3$-Net and ViT degrade close to 2.5\% and 5.0\% on DFDC, respectively. One dominant reason is that these three methods leverage only 1/4 data during fine-tuning, which trades certain performance degradation for faster training and efficient overheads. By contrast, the full fine-tuning methods are more competitive. For example, Face X-ray models significantly surpass Xception on FakeAVCeleb under two settings, and LipForensics outperforms other fine-tuning baselines on DFDC. Finally, our VFD performs the best over all the baselines, \textit{e.g.}, around 12\% improvement over LipForensics. It is worth noting that VFD exploits limited data similar to Xception\_F with only 1/4 dataset, while achieving much satisfactory results. 

\begin{figure}
    \centering
    \includegraphics[width=0.47\textwidth]{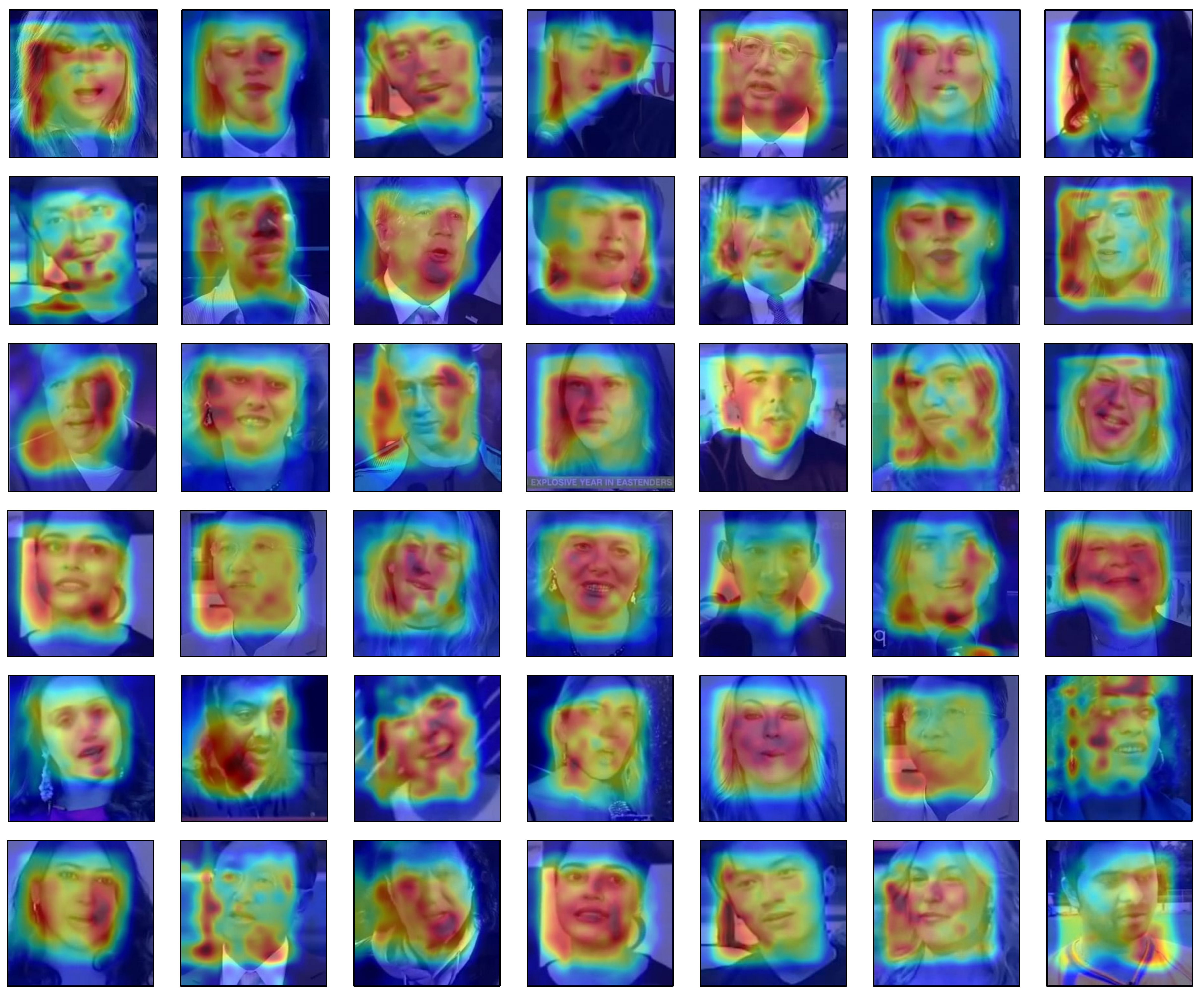}
    \caption{Non-cherry-picked heatmaps of VFD on the FakeAVCeleb dataset}
    \label{extrac_AV}
\end{figure}

\subsection{Qualitative Results}
\subsubsection{Heatmap}
To qualitatively compare our method with baselines, we demonstrated some generated heatmaps. One can observe that in Figure~\ref{same_DFDC} compared with Xception and F$^3$-Net, VFD focuses on the whole face of targets, indicating that VFD recognizes fake videos based on global identity information rather than specific regional artifacts. Moreover, Figure~\ref{extrac_AV} illustrates more VFD heatmaps in FakeAVCeleb. It can be seen that VFD yields stable attention regions on different datasets, which further proves that VFD is more generalizable and will not fail due to the migration of datasets or the update of deepfake algorithms.

\subsubsection{Similarity comparison}
\begin{figure}
    \subfloat[DFDC]{\includegraphics[width=0.24\textwidth]{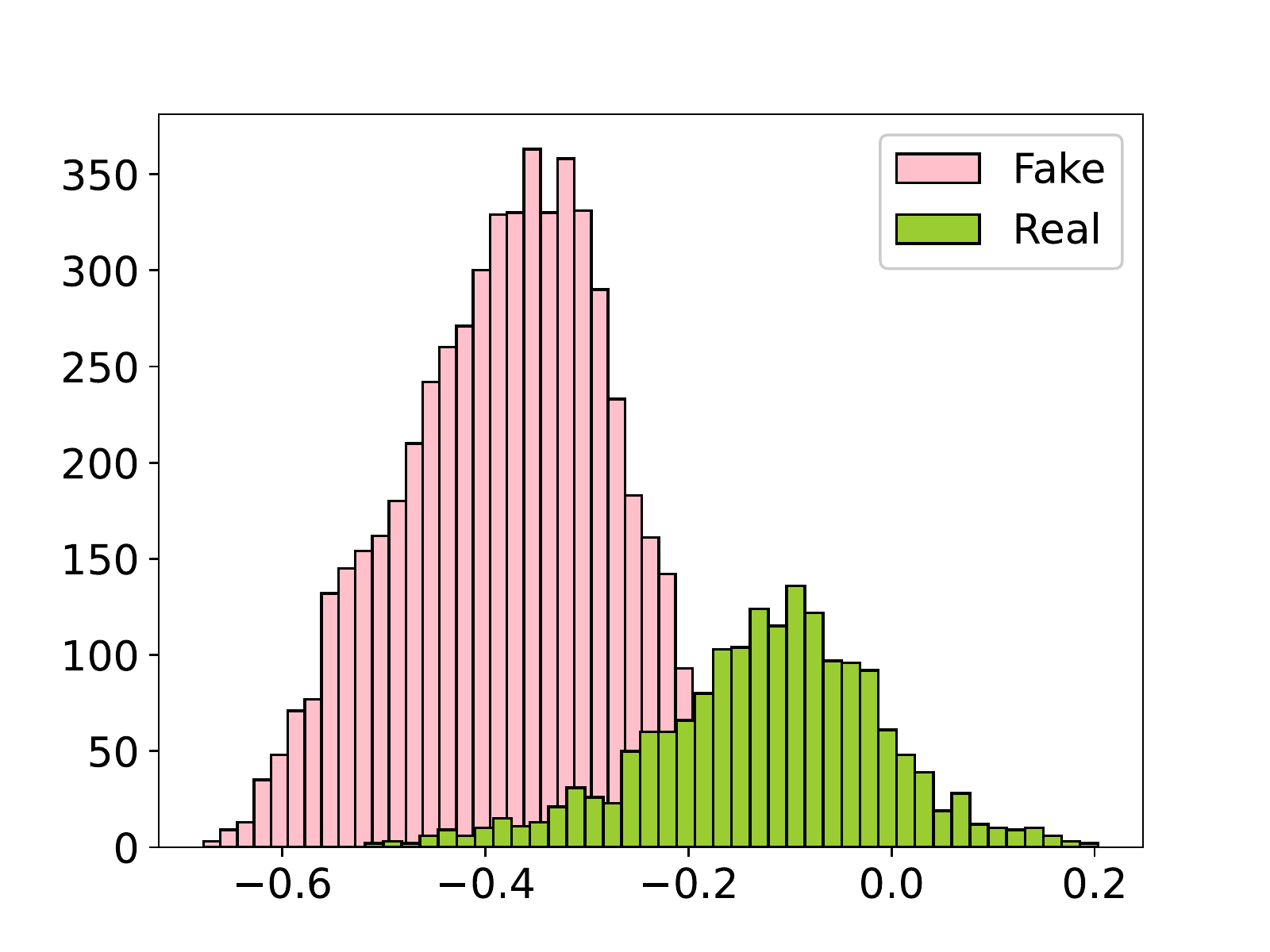}}
    \hfill
    \subfloat[FakeAVCeleb]{\includegraphics[width=0.24\textwidth]{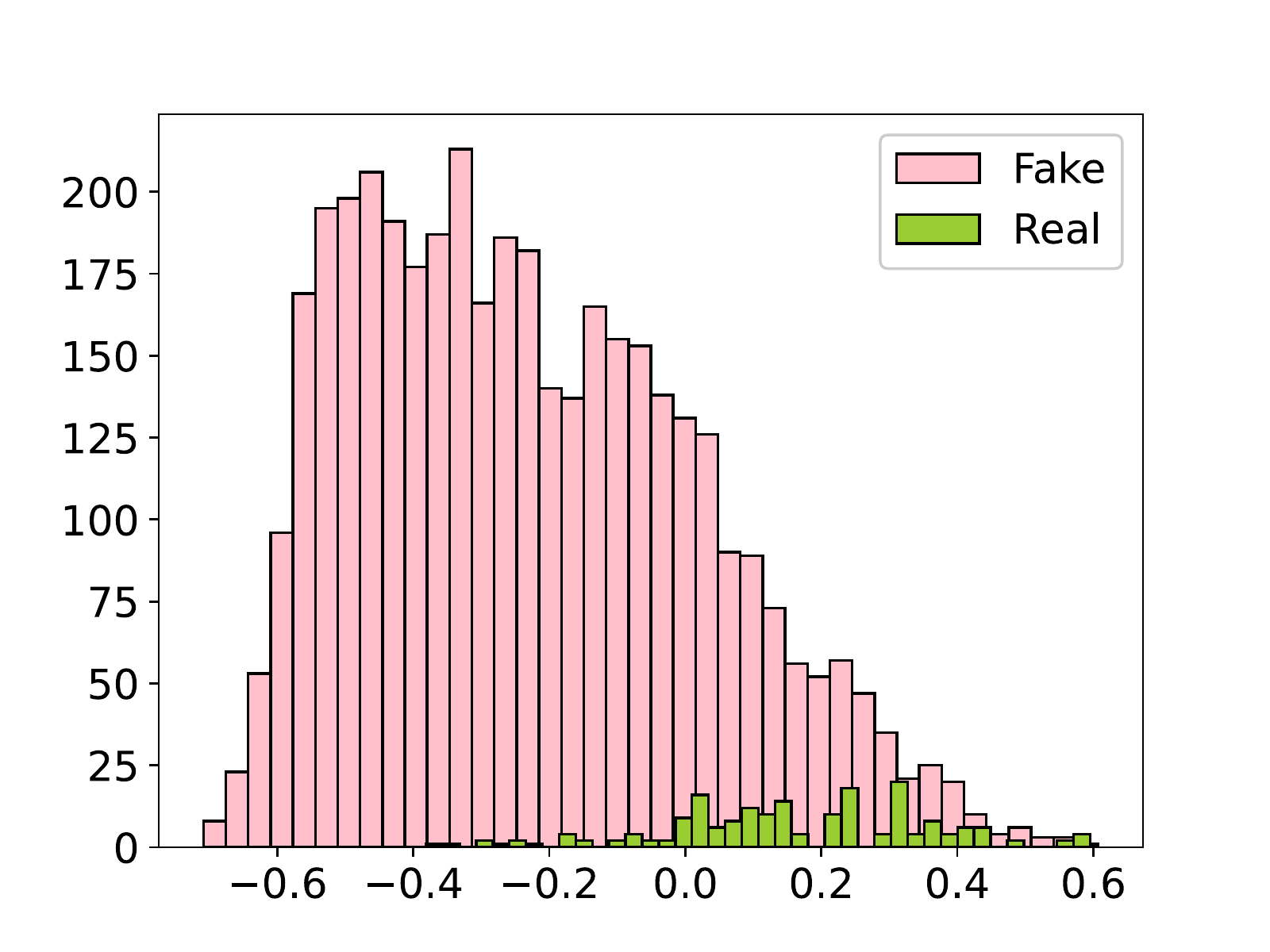}}
    \hfill
    \caption{The similarity between voices and faces. The x-axis represents the similarities, and the y-axis is the number of videos.}
    \label{distance_dataset}
    \vspace{-1em}
\end{figure}
We computed the voice-face similarity from VFD via Equation~\ref{cos_simi} and showed the results in Figure~\ref{distance_dataset}. It demonstrates that the real and fake videos are evidently split. In addition, The split line of real and fake videos is around -0.1 for both datasets, implying that VFD does not calculate the distance via specific features of the dataset but from the general matching view. Otherwise, different cut-off values will be learned by our method.

\begin{figure}
    \subfloat[LQ\_LQ\label{LL}]{\includegraphics[width=0.24\textwidth]{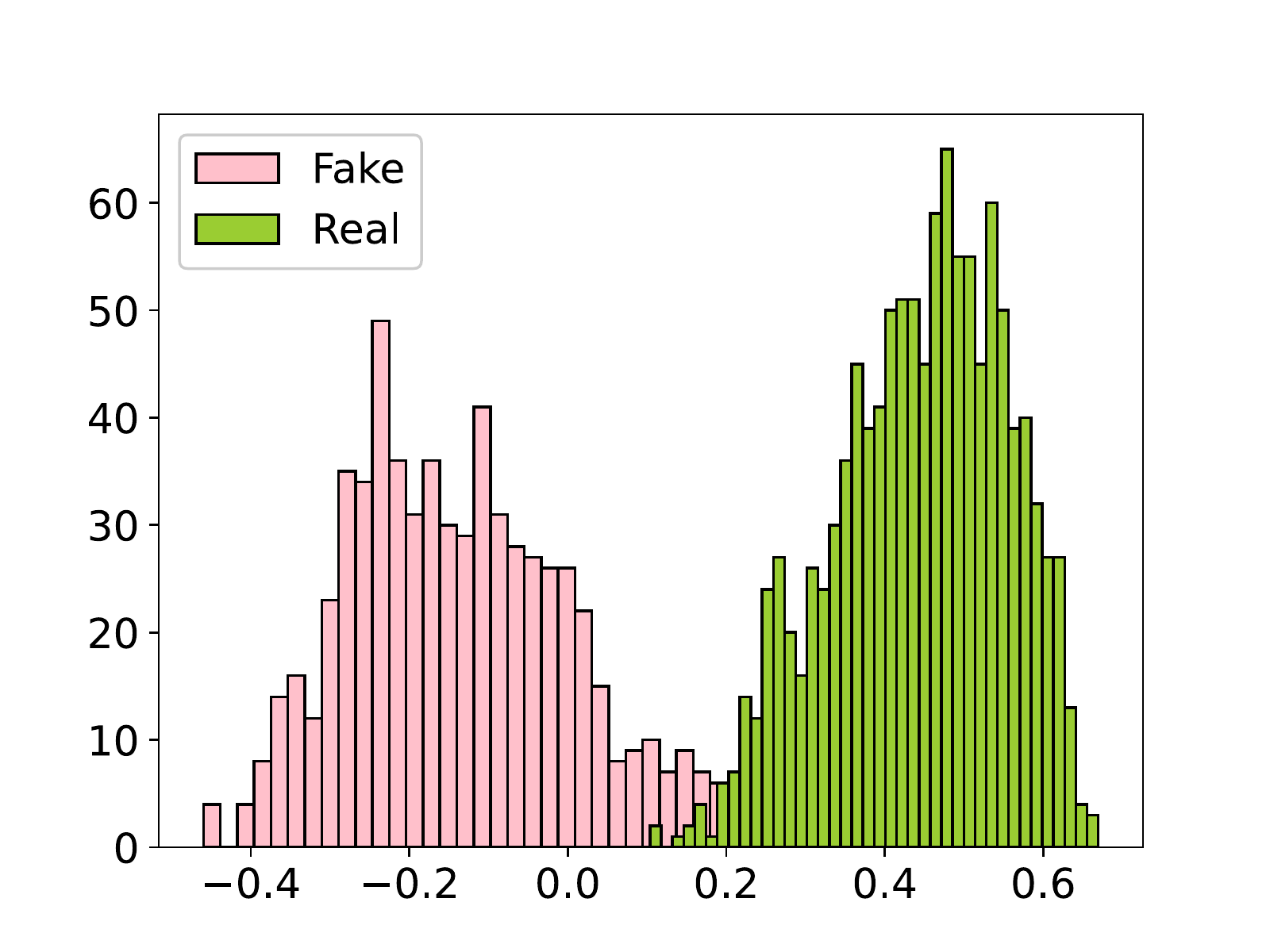}}
    \hfill
    \subfloat[HQ\_HQ\label{HH}]{\includegraphics[width=0.24\textwidth]{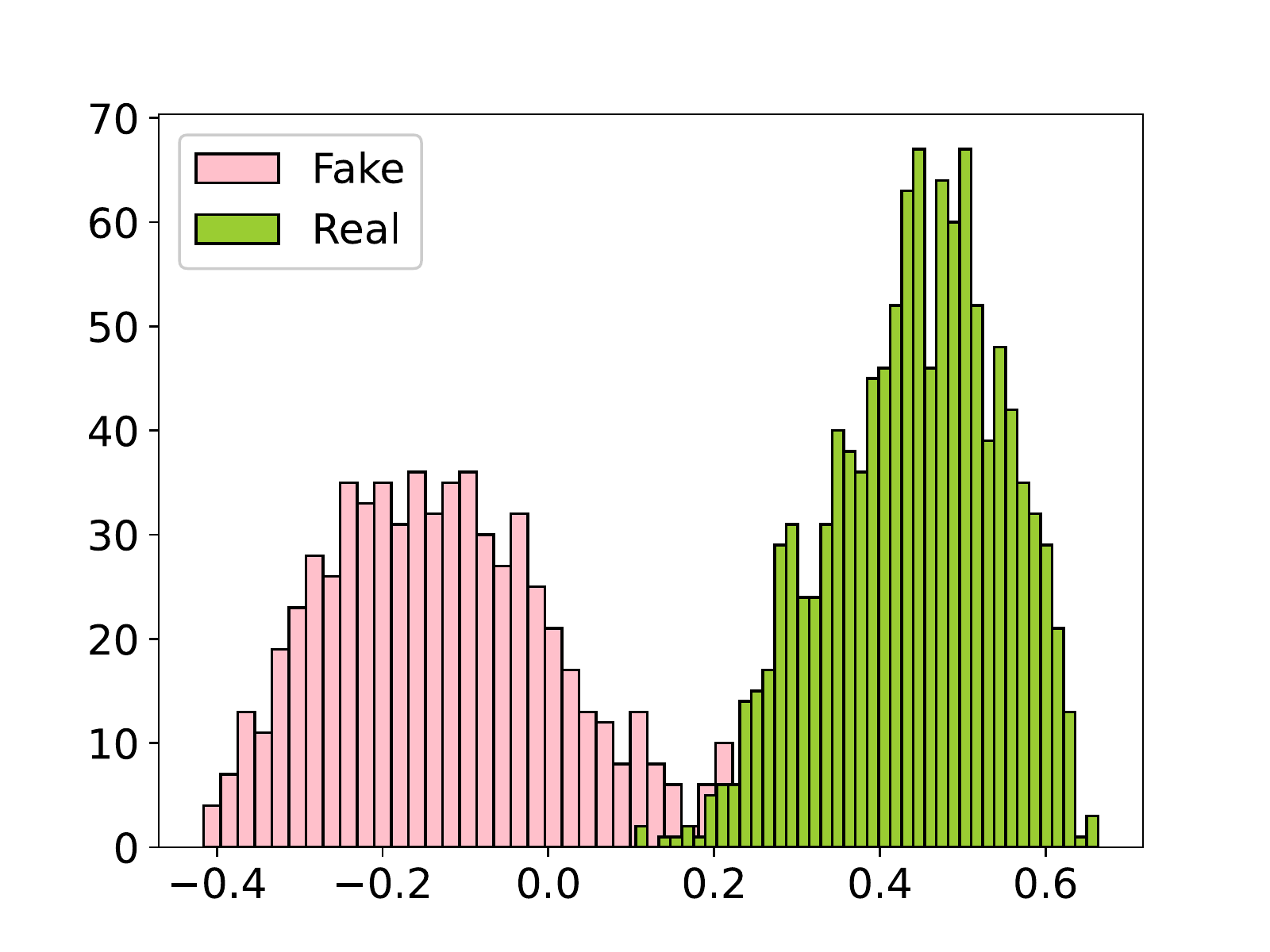}}
    \hfill \\
    \subfloat[LQ\_HQ\label{LH}]{\includegraphics[width=0.24\textwidth]{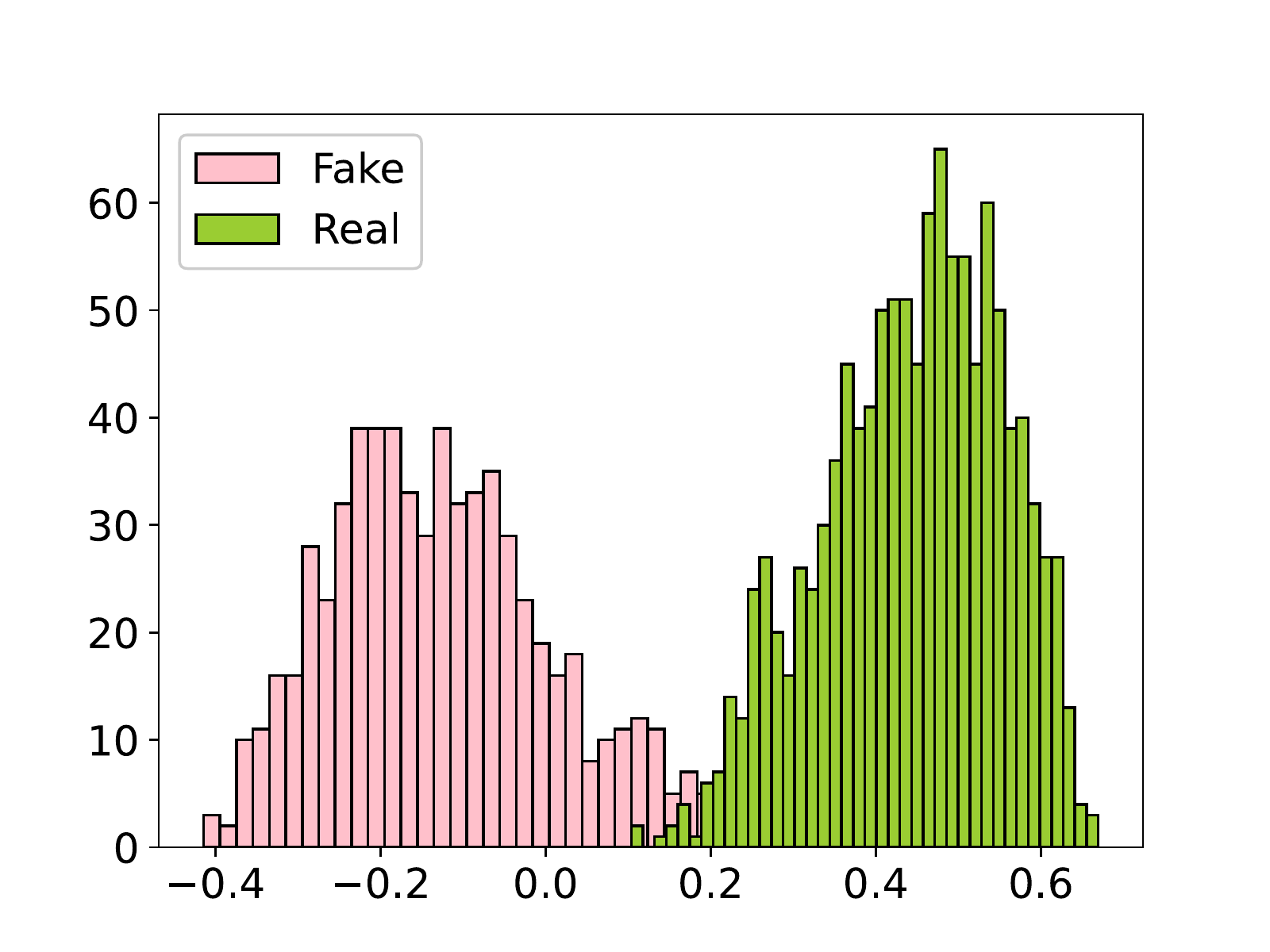}}
    \hfill
    \subfloat[HQ\_LQ\label{HL}]{\includegraphics[width=0.24\textwidth]{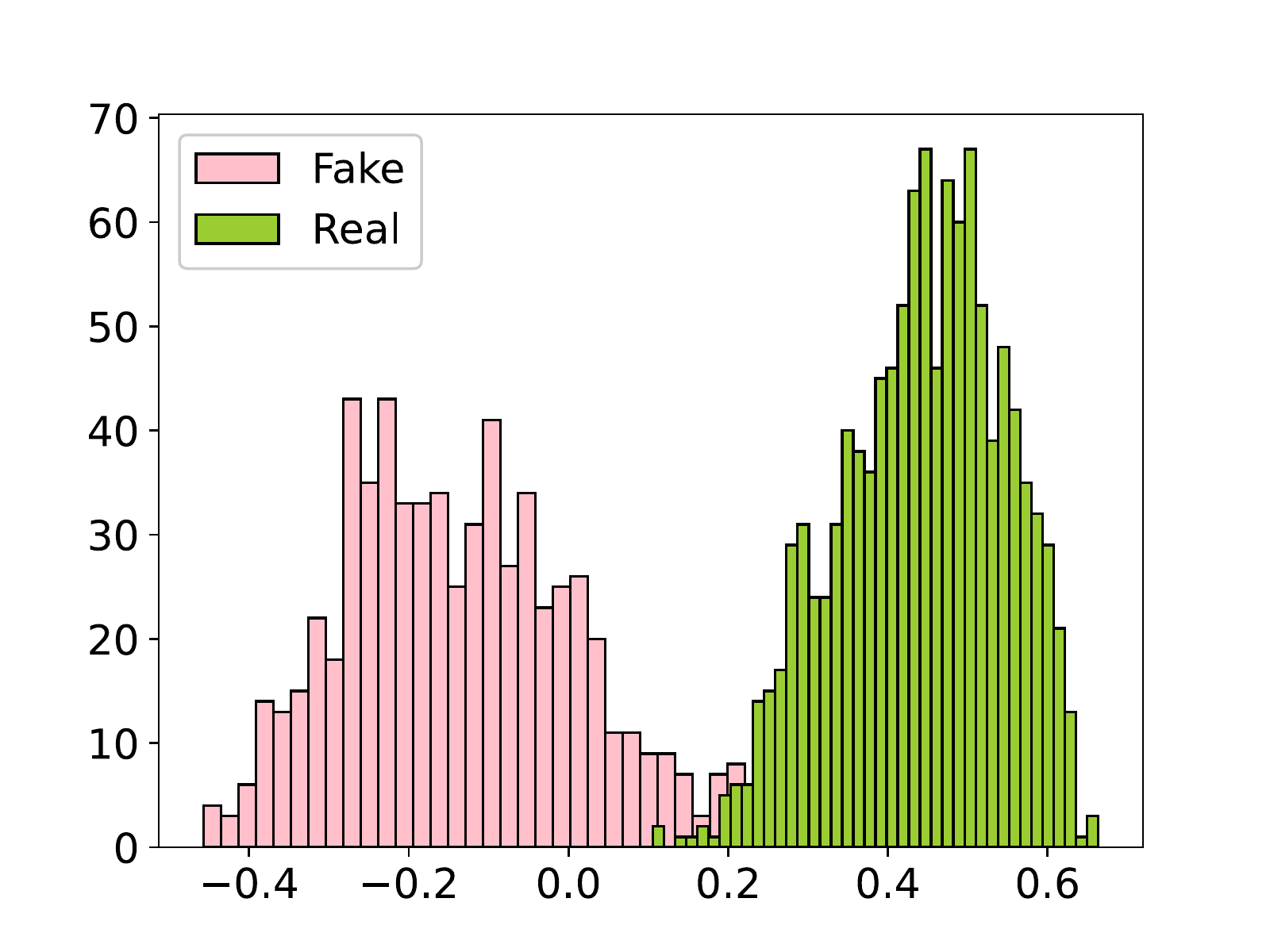}}
    \hfill
    \caption{The similarity between voices and faces in two subset of DeepfakeTIMIT. We captioned the utilized datasets in \{fine-tuning set\}\_\{testing set\} format.}
    \label{TIMIT_Fig}
\end{figure}

Figure~\ref{TIMIT_Fig} illustrates the voice-face similarity on the LQ and HQ subsets of DeepfakeTIMIT. We first show the performance via fine-tuning and testing on LQ and HQ in Figure~\ref{LL} and Figure~\ref{HH}, respectively. It can be observed a clear cut-off line around -0.1, as in DFDC and FakeAVCeleb. Given that the two subsets are synthesized from the same real videos, we swapped fine-tuning sets to verify the reliability of VFD. Specifically, we conducted a new experiment with fine-tuning on HQ and then testing on LQ directly, or vice versa. The results are displayed in Figures~\ref{LH} and Figures~\ref{HL}. It can be seen that the model still discriminates between real and fake videos, proving that the image quality does not significantly affect the effectiveness of the VFD, the dominant reason being that our model focuses on the identity features of the whole face rather than the limited facial manipulation trajectories that are easily covered by compressing or blurring. Another interesting point is that the real parts share a similar distribution when processed via the models fine-tuning on a specific dataset (as in Figure~\ref{LL} and Figure~\ref{LH}). We attribute this phenomenon to the real videos in both LQ and HQ testing sets sampled from Vid-TIMIT. In other words, the real instances in the LQ and HQ sets are identical, thus bringing the same distribution.

\subsubsection{Detection failure cases}
\begin{figure}
    \centering
    \subfloat[Illumination\label{F1}]{\includegraphics[width=0.45\textwidth]{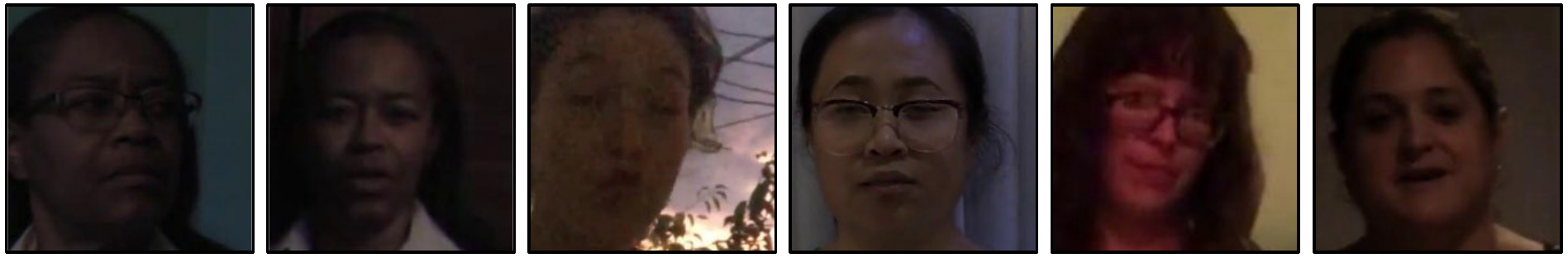}}
    \hfill
    \subfloat[Facial perspectives\label{F2}]{\includegraphics[width=0.45\textwidth]{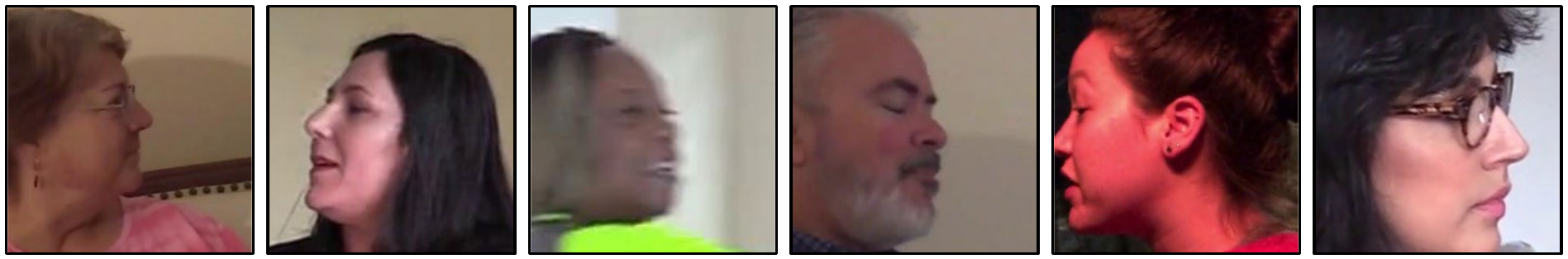}}
    \hfill
    \subfloat[Attributes\label{F3}]{\includegraphics[width=0.45\textwidth]{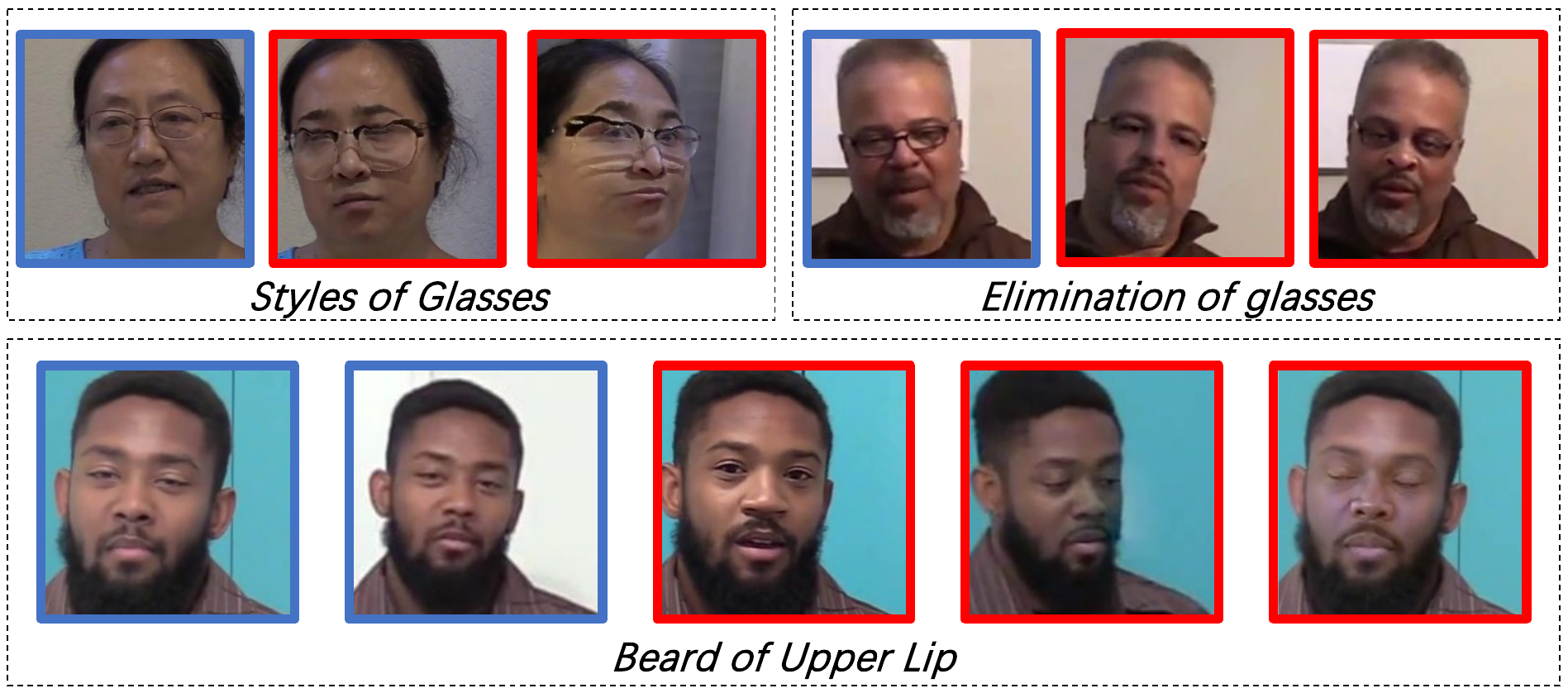}}
\caption{Some detection failure cases. (a) shows blurred faces due to the scarcity of illumination. (b) illustrates extreme facial angles, making the facial features under represented. (c) The modification of facial attributes, \textit{e.g.}, glasses or beard, wherein identities are unchanged.}
\label{fail} 
\end{figure}

We also illustrated the typical failure cases of VFD in Figure~\ref{fail}. There are three notable causes leading to the detection failure, \textit{i.e.}, illumination, facial perspectives, and attributes. From Figure~\ref{F1}, one can see that the faces are mixed with backgrounds due to inadequate illumination, which undermines the model's capability. Figure~\ref{F2} shows some extreme facial angles, \textit{e.g.}, 90 degrees of head-turning, making the facial features difficult to extract. Moreover, some face attribute editing videos~\cite{stylegan} are displayed in Figure~\ref{F3}. The images with a blue border are the faces from real videos in each rectangular box, while red ones are extracted from the corresponding fake videos. One can see that the modification of specific facial attributes, \textit{e.g.}, style of the glasses or upper lip beard, will lead to invalid identification as identities are frozen in this scenario. 

\subsection{Hyperparameter Study}
\begin{figure}
    \subfloat[Number of real videos\label{Finetuning_size}]{\includegraphics[width=0.24\textwidth]{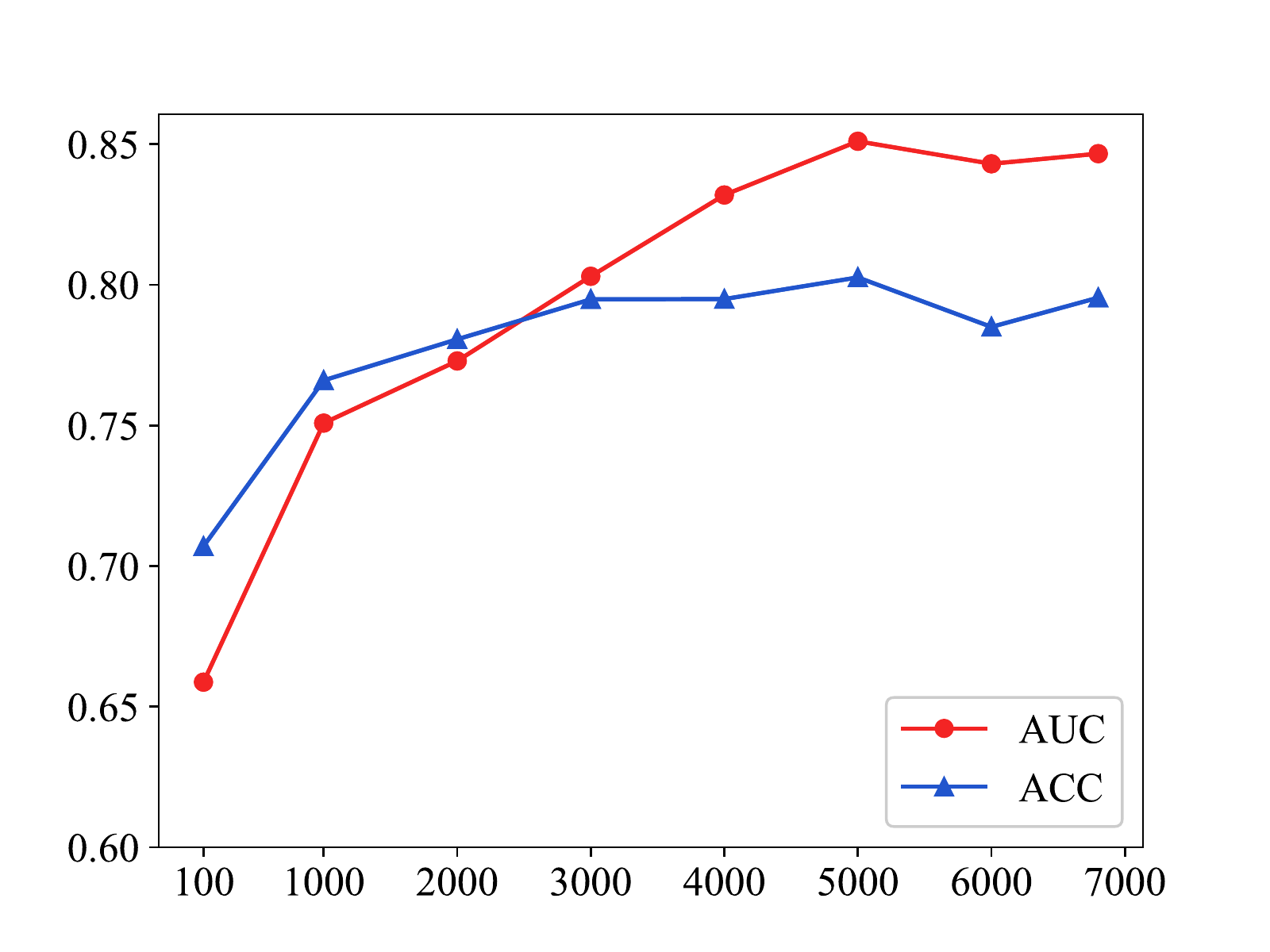}}
    \hfill
    \subfloat[Number of negatives\label{QChoice}]{\includegraphics[width=0.24\textwidth]{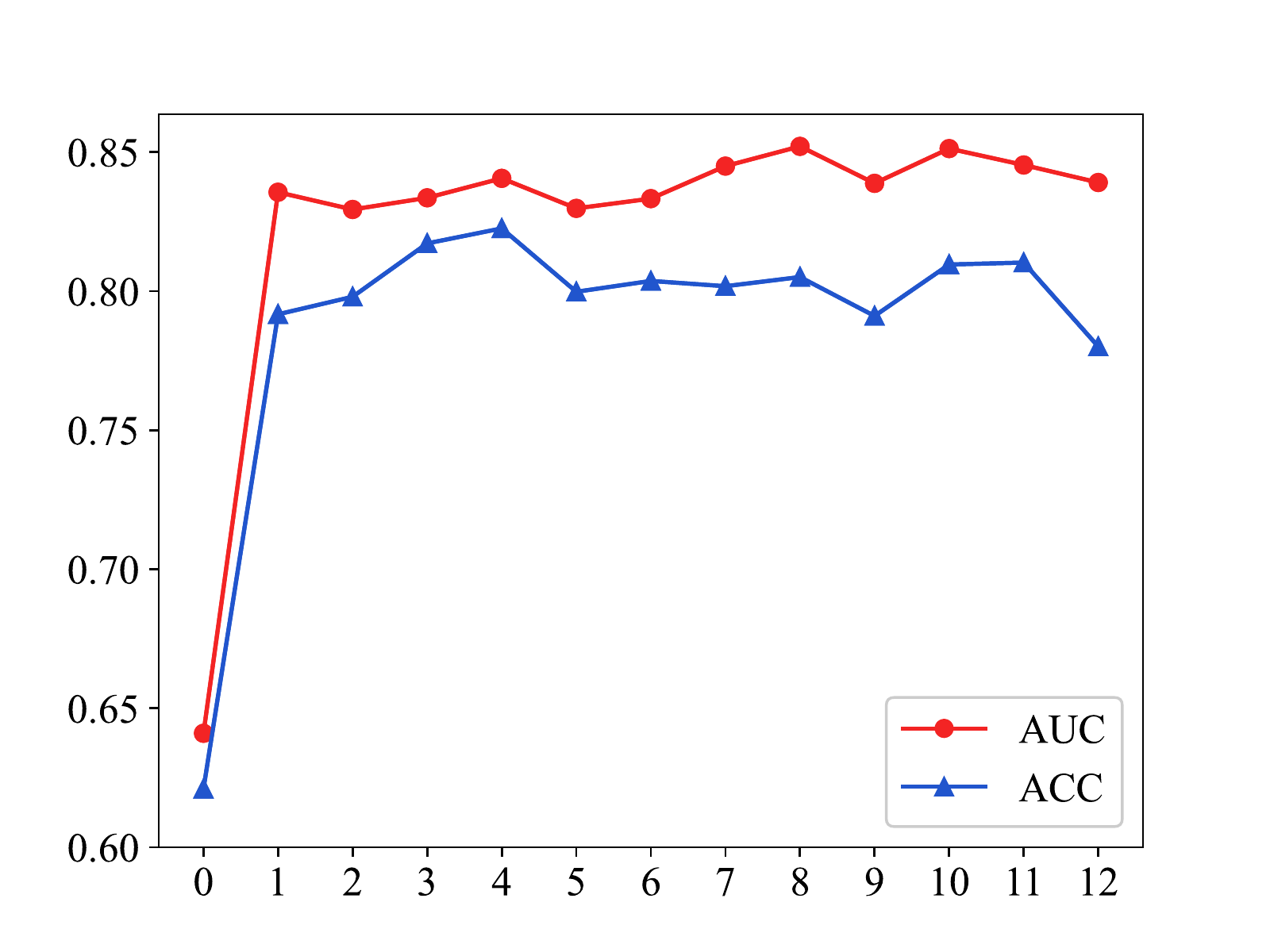}}
    \hfill
    \caption{Hyperparameter study. (a) The AUC and ACC results w.r.t. different numbers of real videos in the deepfake dataset. (b) The influence of the negative samples.}
    \label{parameter_testing}
\end{figure}

We studied VFD with extensive parameter settings. We first illustrated the AUC and ACC results with different fine-tuning dataset sizes on DFDC in Figure~\ref{Finetuning_size}. As we can see, when fine-tuning on a limited subset of fine-tuning dataset (\textit{e.g.}, 1,000 or 2,000 real samples), VFD already achieves an AUC of 77\%, which outperforms most baselines. Furthermore, even if the dataset degrades to 100, VFD's performance (65.87\% on AUC) is still comparable to some baselines such as MesoInception-4 (60.56\%) and VA-LogReg (66.20\%), proving that VFD can achieve an acceptable identification ability in a few-shot way and quickly adapt to newly forgery algorithms. VFD reaches its best performance when fine-tuning instances reach 5,000. Compared to traditional approaches where the entire DFDC training set is used, \textit{i.e.}, more than 20,000 videos~\cite{F3Net,dfd1,dfd4}, VFD requires fewer samples to achieve better performance, making our method distinguished from these baselines.

In addition, we also studied the influence of different number of negatives, \textit{i.e.}, the $\mathrm{Q}$ in Equation~\ref{RFD}, and show the results in Figure~\ref{QChoice}, The number of 0 denotes no fake pairs are leveraged for fine-tuning. We can observe that the performance is significantly boosted with more negatives being taken into the model. When the number reaches 10, the model performance saturates and benefits no further gains. 

\begin{table*}
\centering
\caption{Performance of VFD under different ablation settings on DFDC and FakeAVCeleb.}
\begin{tabular}{lccccccccc}
\toprule
\multicolumn{1}{c}{\multirow{2}{*}{Model}} & \multicolumn{2}{c}{Modality} & \multirow{2}{*}{Modulator} & \multirow{2}{*}{InfoNCE} & \multirow{2}{*}{RFC} & \multicolumn{2}{c}{DFDC} & \multicolumn{2}{c}{FakeAVCeleb} \\ \cmidrule(lr){2-3} \cmidrule(lr){7-8} \cmidrule(lr){9-10} 
                       & Visual        & Audio        &                            &                    &             & AUC         & ACC        & AUC            & ACC      \\ \midrule
Visual-only                     & \checkmark    &              &                            &                    &             &65.04 &71.16 &70.05 & 74.30                \\
Visual-auditory                     & \checkmark    & \checkmark   &                            &                    &             &52.45 &54.52 &64.16 & 69.12               \\
Modulator                     & \checkmark    & \checkmark   & \checkmark                 &                    &             &56.93 &59.54 &65.01 & 69.22                \\
InfoNCE-based                     & \checkmark    & \checkmark   & \checkmark                 &\checkmark          &             &59.64 &60.15 &80.15 & 85.40                \\ \midrule
VFD                    & \checkmark    & \checkmark   & \checkmark                 &\checkmark          &\checkmark  &\textbf{80.96}  &\textbf{85.13} & \textbf{81.52}  &\textbf{86.11}\\   \bottomrule  
\end{tabular}
\label{ablation}
\end{table*}

\subsection{Ablation Study}
\label{abl}
\subsubsection{Module evaluation}
To study the effectiveness of different modules in VFD, we explored the performance of the following variants: 1) Visual-only is the visual-modality only version of VFD, wherein a cross-entropy loss is employed to classify the real and fake videos, namely, the ViT\_F in Table~\ref{performance}. 2) Visual-auditory utilizes both voice and face extractors, while the two modulators in Figure~\ref{MMN} are removed, and a linear classifier is applied to bind multi-modal features instead of contrastive loss. 3) Modulator adds the voice and face modulators to M1. 4) InfoNCE-based model integrates InfoNCE into the pre-training phase while replacing the proposed RFC with InfoNCE in the fine-tuning stage.

As can be seen from Table~\ref{ablation}, all these four variants will jeopardize the performance to some extent. Among them, the Visual-auditory and Modulator models perform inferior to the uni-modal counterpart Visual-only in most cases, proving that detecting deepfakes via multi-modal cues require elaborated approaches to model the homogeneity. Furthermore, Modulator surpasses than Visual-auditory due to the introduction of modality modulators. Moreover, it can be observed that applying InfoNCE to pre-training will improve the model performance, validating the effectiveness of our pre-training strategy. Meanwhile, the InfoNCE-based model performs comparably with RFC on FakeAVCeleb while fails severely on DFDC during fine-tuning. We attributed this result to the fact that the FakeAVCeleb is built with relatively detailed annotation, \textit{i.e.}, both voice and face manipulations are labeled, which enables InfoNCE to obtain adequate features. Nevertheless, when a similar strategy is applied to coarsely annotated datasets, \textit{e.g.}, DFDC, some crucial properties are overlooked, as introduced in Section~\ref{finetuning}. In a nutshell, when combining all the devised modules, our method can achieve the best result.

\subsubsection{Strategy evaluation}
To further demonstrate the necessity of the pre-training then fine-tuning paradigm, we studied the following variants: 1) VFD w/o pre-training replaces the pre-training strategy with training from scratch. 2) VFD w/o fine-tuning applies the pre-trained model to detect forgery directly. From Table~\ref{ablation_2}, one can observe that the pre-training strategy contributes more than 10\% and 5\% AUC improvements on DFDC and FakeAVCeleb, respectively, wherein the homogeneity between voices and faces are accordingly modeled. Meanwhile, the removal of fine-tuning is more detrimental to the performance, \textit{e.g.}, the AUC on DFDC drops by 30\%, indicating that fine-tuning is essential to narrow the gap between generic and deepfake datasets.

\begin{table}[t]
\centering
\begin{center}
\caption{Performance of VFD when disabling pre-training or fine-tuning.}
\label{ablation_2}
\begin{tabular}{lcccc}
\toprule
\multicolumn{1}{c}{\multirow{2}{*}{Model}}  & \multicolumn{2}{c}{DFDC}    & \multicolumn{2}{c}{FakeAVCeleb} \\ \cmidrule(lr){2-3} \cmidrule(lr){4-5}
\multicolumn{1}{c}{}     & ACC               & AUC             & ACC      & AUC    \\ \midrule
VFD w/o pre-training &73.05 &74.79 &75.30 & 79.53 \\
VFD w/o fine-tuning &50.61 &53.40 &72.76 & 76.05 \\ \midrule
VFD   & {\textbf{80.96}}  & \textbf{85.13}   & \textbf{{81.52}}       & \textbf{86.11} \\ 
\bottomrule
\end{tabular}
\end{center}
\end{table}

\subsubsection{Feature extractor evaluation}
We investigated the impact of feature extractors structures. Specifically, we modified the transformer-based models in Section~\ref{pretraining} into a VGG-based~\cite{VGG_based} ones. We then evaluated their performance in two dimensions, \textit{i.e.}, pre-training and fine-tuning. The pre-training reports the capability of the each feature extractor for identifying the faces and voices on the generic dataset, and the accuracy on matching voice-face pairs from the same identities. Besides, fine-tuning focuses on the performance on the deepfake dataset, we thus compare the AUC and ACC results on DFDC dataset.

\begin{table}[t]
\centering
\begin{center}
\caption{Performance of different extractors on pre-training and fine-tuning.}
\label{ablation_3}
\begin{tabular}{lccccc}
\toprule
\multicolumn{1}{c}{\multirow{2}{*}{Model}}  & \multicolumn{3}{c}{Pre-training}    & \multicolumn{2}{c}{Fine-tuning} \\ \cmidrule(lr){2-4} \cmidrule(lr){5-6}
\multicolumn{1}{c}{}     & Face       & Voice    & Match          & ACC      & AUC    \\ \midrule
VGG-based                 & 94.67   & 72.62    & 81.54            & 72.19 & 75.57 \\
Transformer-based (VFD)   & {\textbf{98.62}}   & \textbf{87.25}   & \textbf{{96.90}}     & \textbf{80.96} & \textbf{85.13}\\ 
\bottomrule
\end{tabular}
\end{center}
\end{table}

From Table~\ref{ablation_3}, one can observe that The transformer-based model outperforms the VGG-based model in all metrics in this task. We attribute this to the transformers' superiority to model the global features of faces and voices, which will be more suitable for identity-based matching. Besides, both types of extractors show superior performance in face recognition, \textit{i.e.}, over 98\% and 94\% accuracy in classifying the faces, respectively. The identification of voice is the bottleneck that limits the model performance, implying that accurate classification for voice is both challenging and cutting-edge. Moreover, VFD exhibits a matching accuracy close to 97\% on the generic dataset. A possible explanation is that the multi-modal inputs are mutually supportive via contrastive learning, such that the errors of the voice feature extractor are compensated to some extent. Meanwhile, the decrease after migration to the deepfake dataset suggests that our matching view has considerable potential.

%% file: 5_conclusion.tex
\section{Conclusion and Discussion}
\label{Sec:con}
Detecting forgery videos in deepfake is challenging due to the continuous progress from deepfake techniques. In this work, we have empirically recognized the severe mismatch of the voices and faces in deepfake videos, based on which we proposed to tackle this task from an unexplored homogeneity modeling perspective. In particular, our proposed method follow a pre-training then fine-tuning pipeline, wherein the voices and faces from a single identity are learned to closely match with a novel RFC loss. The experimental and visualization results demonstrate the effectiveness of the proposed method against existing elaborately designed competitors.

Despite its effectiveness on existing deepfake datasets, VFD shows a certain limitation in the unusual case where both voices and faces are maliciously edited. Given the fact that our method addresses deepfake detection from either face swapping or voice editing, simultaneously performing these two may lead to the matching collapse. However, to the best of our knowledge, no such high-quality datasets, as well as associated approaches, have been studied in literature so far. Yet, this challenging setting is still promising and demands extensive exploration in the future.